\documentclass{article}

\usepackage{arxiv}
\usepackage[utf8]{inputenc} 
\usepackage[T1]{fontenc}  
\usepackage[acronym]{glossaries}
\usepackage{hyperref}       
\usepackage{url}            
\usepackage{booktabs}       
\usepackage{amsfonts}       
\usepackage{nicefrac}       
\usepackage{microtype}      
\usepackage{lipsum}		
\usepackage{adjustbox}
\usepackage{graphicx}
\usepackage{natbib}
\usepackage{doi}
\usepackage{listings}
\usepackage{subcaption}
\usepackage{xcolor}
\usepackage{graphicx}
\usepackage{multirow}
\usepackage{fancyhdr}
\lstdefinestyle{pytorch}{
    language=Python,
    backgroundcolor=\color{white},
    commentstyle=\color{gray},
    keywordstyle=\color{blue},
    stringstyle=\color{red},
    basicstyle=\ttfamily,
    breaklines=true,
    frame=single,
}
\usepackage{authblk} 
\date{}
\title{Seeing Eye to AI: Human Alignment via Gaze-Based Response Rewards for Large Language Models. \thanks{This paper has been accepted to ICLR 2025. The final version will be published in the conference proceedings.}}

\author[1,2]{Angela Lopez-Cardona}
\author[1]{Carlos Segura}
\author[3]{Alexandros Karatzoglou}
\author[2]{Sergi Abadal}
\author[1]{Ioannis Arapakis}

\affil[1]{Telefónica Scientific Research, Barcelona, Spain}
\affil[2]{Universitat Politècnica de Catalunya, Barcelona, Spain}
\affil[3]{Amazon, Barcelona, Spain}

\affil{\texttt{angela.lopezcardona@telefonica.com, carlos.seguraperales@telefonica.com, alexandros.karatzoglou@gmail.com, abadal@ac.upc.edu, arapakis.ioannis@gmail.com}}

\makeglossaries
\newacronym{ai}{AI}{Artificial Intelligence}
\newacronym{bilstm}{BiLSTM}{Bidirectional Long Short-Term Memory}
\newacronym{cot}{CoT}{Chain of Thought}
\newacronym{ct}{CT}{Conversation Tree}
\newacronym{cnn}{CNN}{Convolutional Neural Network}
\newacronym{drl}{DRL}{Deep Reinforcement Learning}
\newacronym{ddpg}{DDPG}{Deep Deterministic Policy Gradient}
\newacronym{dl}{DL}{Deep Learning}
\newacronym{dnn}{DNN}{Deep Neural Networks}
\newacronym{dqn}{DQN}{Deep Q-learning}
\newacronym{dpo}{DPO}{Direct Preference Optimization}
\newacronym{ddqn}{DDQN}{Double Q-learning}
\newacronym{et}{ET}{Eye-tracking}
\newacronym{ft}{FT}{Fine-Tuning}
\newacronym{gae}{GAE}{Generalized Advantage Estimate}
\newacronym{gnn}{GNN}{Graph Neural Networks}
\newacronym{gqa}{GQA}{Grouped Query Attention}
\newacronym{hrlaif}{HRLAIF}{Hybrid Reinforcement Learning from AI Feedback}
\newacronym{il}{IL}{Imitation Learning}
\newacronym{ipopt}{IPOPT}{Interior Point Optimizer}
\newacronym{kl}{KL}{Kullback–Leibler}
\newacronym{llm}{LLM}{Large Language Model}
\newacronym{llms}{LLMs}{Large Language Models}
\newacronym{lstm}{LSTM}{Long short-term memory}
\newacronym{lora}{LoRA}{Low-Rank Adaptation}
\newacronym{lr}{LR}{Learning Rate}
\newacronym{mdp}{MDP}{Markov Decision Process}
\newacronym{mha}{MHA}{multi-head attention}
\newacronym{ml}{ML}{Machine Learning}
\newacronym{mlp}{MLP}{Multilayer Perceptron}
\newacronym{mpnn}{MPNN}{Message Passing Neural Networks}
\newacronym{mtl}{MTL}{Multi-task learning}
\newacronym{mse}{MSE}{mean squared error}
\newacronym{mae}{MAE}{mean absolute error}

\newacronym{nlp}{NLP}{Natural Language Processing}
\newacronym{nn}{NN}{Neural Networks}
\newacronym{ocr}{OCR}{Optical Character Recognition}
\newacronym{orms}{ORMs}{Outcome-supervised reward models}
\newacronym{p3o}{P3O}{Pairwise Proximal Policy Optimization}
\newacronym{peft}{PEFT}{Parameter-Efficient Fine-Tuning}
\newacronym{pi}{PI}{Policy Iteration}
\newacronym{ppo}{PPO}{Proximal Policy Optimization}
\newacronym{prms}{PRMs}{Process Based Reward Models}
\newacronym{raft}{RAFT}{Reward rAnked FineTuning}
\newacronym{rag}{RAG}{Retrieval-Augmented Generation}
\newacronym{rnn}{RNN}{Recurrent Neural Network}
\newacronym{rnns}{RNNs}{Recurrent Neural Networks}
\newacronym{rl}{RL}{Reinforcement Learning}
\newacronym{rlaif}{RLAIF}{Reinforcement Learning from AI Feedback}
\newacronym{rlcd}{RLCD}{Reinforcement Learning from Contrastive Distillation}
\newacronym{rlhf}{RLHF}{Reinforcement Learning from Human Feedback}
\newacronym{rm}{RM}{Reward Model}
\newacronym{rso}{RSO}{Statistical Rejection Sampling Optimization}
\newacronym{rrhf}{RRHF}{Rank Responses to align Human Feedback}

\newacronym{sota}{SOTA}{State of the Art}
\newacronym{sft}{SFT}{Supervised Fine-Tuning}
\newacronym{smoe}{SMoE}{Sparse Mixture of Experts}
\newacronym{slic}{SLiC}{Sequence Likelihood Calibration}
\newacronym{slcihf}{SLiC-HF}{Sequence Likelihood Calibration with Human Feedback}
\newacronym{sorms}{SORMs}{Stepwise ORMs}
\newacronym{swa}{SWA}{sliding window attention}
\newacronym{trpo}{TRPO}{Trust Region Policy Optimization}
\newacronym{trt}{TRT}{total reading time}

\newacronym{vi}{VI}{Value Iteration}

\begin{document}

\maketitle
\begin{abstract}
Advancements in Natural Language Processing (NLP), have led to the emergence of Large Language Models (LLMs) such as GPT, Llama, Claude, and Gemini, which excel across a range of tasks but require extensive fine-tuning to align their outputs with human expectations. A widely used method for achieving this alignment is Reinforcement Learning from Human Feedback (RLHF), which, despite its success, faces challenges in accurately modelling human preferences. In this paper, we introduce GazeReward, a novel framework that integrates implicit feedback -- and specifically eye-tracking (ET) data -- into the Reward Model (RM). In addition, we explore how ET-based features can provide insights into user preferences. Through ablation studies we test our framework with different integration methods, LLMs, and ET generator models, demonstrating that our approach significantly improves the accuracy of the RM on established human preference datasets. This work advances the ongoing discussion on optimizing AI alignment with human values, exploring the potential of cognitive data for shaping future NLP research.
\end{abstract}

\section{Introduction}

Recent advancements in \acrfull{nlp} have led to the emergence of \acrfull{llms} like GPT \citep{openai_gpt-4_2023}, Llama \citep{dubey_llama_2024}, Claude \citep{anthropic_claude_2024-1}, and Gemini \citep{gemini_team_gemini_2024}, which excel across a range of tasks.
These models, often consisting of billions of parameters, are trained on massive datasets and typically require extensive fine-tuning to align their outputs with human expectations \footnote{\acrshort{llms} that are trained only on extensive datasets for language modeling are referred to as ``pre-trained'' \acrshort{llms}. Subsequent approaches, such as human alignment, are categorized as ``post-training''.}. Several 
works have focused on refining the way \acrshort{llms} interpret and respond to user intent \citep{wang_aligning_2023}, which has led to the development of novel alignment techniques. 
A common approach to achieving human alignment involves leveraging explicit human feedback as preference information. Currently, the most widely adopted method is \acrfull{rlhf} \citep{ouyang_training_2024}. \acrshort{rlhf} has been implemented in many state-of-the-art \acrshort{llms} \citep{cui_ultrafeedback_2024, openai_gpt-4_2023, bai_constitutional_2022}, and has been shown to help align models to human instructions and mitigate the generation of toxic or harmful content \citep{kiegeland_pupil_2024}. However, a persistent challenge with this approach is the difficulty of acquiring sufficient high-quality training data \citep{casper_open_2023}.

To be able to capture the complexities of real-world user instructions, there is a need for meticulously handcrafted data \citep{wang_aligning_2023}, which are resource-expensive and difficult to scale \citep{yang_harnessing_2023}. Obtaining high-quality feedback from human annotators, usually provided after examining a model response, suffers from several caveats \citep{casper_open_2023}. For instance, low inter-annotator agreement can result in inconsistent evaluations of the same model output due to varying interpretations, domain expertise, or biases. Moreover, ``scalable oversight'' -- the ability to supervise models effectively with limited resources \citep{amodei_concrete_2016} -- remains an open problem. Inconsistent data quality is another issue, as cost-quality tradeoffs often arise when collecting human feedback. 



To address these challenges, researchers have 
increasingly turned to \acrshort{llms} as a form of AI-driven feedback, referred to as \acrfull{rlaif} \cite{bai_constitutional_2022}. This method offers improved scalability, easier data collection, and cost-efficiency compared to traditional human-driven approaches \citep{bai_constitutional_2022, wang_self-instruct_2023, madaan_self-refine_2023}. However, it remains unclear what type of feedback signals, or a combination of feedback mechanisms, is optimal to align \acrshort{llm} with human goals \citep{casper_open_2023}. More research is needed to explore the underlying beliefs and expectations of human users \citep{casper_open_2023}, and how these can be incorporated into human alignment techniques. Furthermore, the alignment success of a language model is dependent on the quality of the underlying \acrshort{rm} \citep{pace_west--n_2024}. Various alignment methods, such as \acrshort{rlhf}, \acrshort{rlaif}, and \acrfull{dpo} \citep{rafailov_direct_2023}, 
rely on \acrshort{rm} to incorporate feedback. Reward modelling is also essential for generating synthetic data for preference alignment and is often used in \acrshort{llm} inference to evaluate model outputs in techniques such as best-of-N sampling \citep{cui_ultrafeedback_2024}.

In this work, we propose a novel approach that incorporates \acrfull{et} as an additional signal to address the challenge of human alignment. \acrshort{et} measures oculomotor behavior i.e. the movements and fixations of the eyes, which offers insight into visual attention and information processing \citep{kleinke_gaze_1986, land_knowledge_1997}. This allows researchers to correlate observable eye movement patterns with underlying cognitive and perceptual processes during reading and language comprehension tasks \citep{kleinke_gaze_1986, krasich_gaze-based_2018}. Moreover, \acrshort{et} -- unlike other (explicit) forms of feedback (e.g., questionnaire data, data annotation) -- does not suffer from human biases, and offers a better temporal and spatial resolution \citep{zhang_eye-tracking_2024}. 
Several studies have shown a strong correlation between human eye movements and attention patterns in transformer-based models \citep{wang_gaze-infused_2024, bensemann_eye_2022, sood_interpreting_2020}. Incorporating \acrshort{et} data into \acrshort{nlp} tasks has also proven valuable, as demonstrated by numerous works \citep{huang_longer_2023, khurana_synthesizing_2023, hollenstein_advancing_2019, yang_plm-as_2023,kiegeland_pupil_2024, deng_pre-trained_2023, mathias_eyes_2018, mcguire_sentiment_2021}. Recently, \cite{kiegeland_pupil_2024} proposed the integration of \acrshort{et} in controlled sentiment generation to create a dataset that can be used in human alignment methods.

\begin{figure}[h]
    \begin{center}
    \includegraphics[width=\linewidth, trim=2mm 2mm 2mm 2mm, clip]{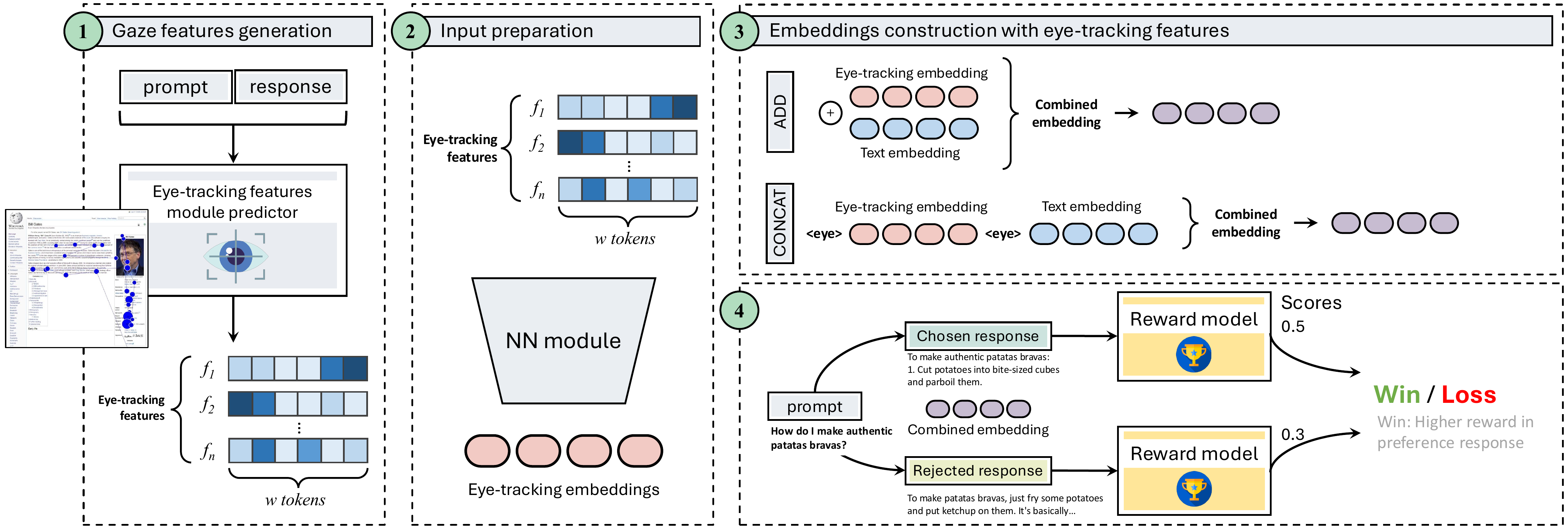}
    \end{center}
    \caption{GazeReward Framework for using eye-tracking data for reward modelling. We use a generator model to compute \acrshort{et} features on a preference dataset $D$ and we train the human preference by combining both text and \acrshort{et} embeddings (See \autoref{sec:method} for details.)}
    \label{fig:pipeline}
\end{figure}

Human alignment remains one of the biggest challenges in the development of \acrshort{llm}, with \acrshort{rm} playing an important role in addressing this issue. This paper investigates how behavioural signals, particularly \acrshort{et}, can be operationalised as implicit feedback to improve human alignment (proposed approach is shown in \autoref{fig:pipeline}). Furthermore, we explore the use of \acrshort{et} prediction models that can generate -- automatically and with little effort -- \acrshort{et} features in response to text input, which makes our solution not only cost-effective but also highly scalable. 

Our contributions are the following:

\begin{itemize}
    \item We propose \textbf{GazeReward}, a novel and scalable framework that integrates implicit feedback in the form of \acrshort{et} data into the \acrshort{rm}, a key component in modeling human preferences.
    \item We perform for the first time an ablation study that examines several state-of-the-art \acrshort{llms}, various \acrshort{et} prediction models, and methods for incorporating \acrshort{et} features into the \acrshort{rm}.
    \item We demonstrate experimentally substantial performance improvements with the GazeReward  framework, showing accuracy gains of over 10\% in \acrshort{rm} predictions across diverse human preference datasets.
\end{itemize}

\section{Preliminaries}
\subsection{\acrlong{llms}-Human Alignment}
\acrshort{llms}-Human Alignment typically involves training \acrshort{llms} \footnote{Before the process of human alignment, these models are referred to as ``pre-trained'' LLMs.} on datasets curated by humans (learning from human feedback data) \citep{ouyang_training_2024}. This can be achieved through \acrfull{sft}, where the model is trained on pairs of prompts ($x$) and corresponding human-generated responses ($y$) \citep{liu_statistical_2024}. Alternatively, alignment can be pursued via preference optimization, using a human preference dataset that differentiates between a better response ($y_w$) and a worse one ($y_l$) for the same prompt ($x$): $\mathcal{D} = \{(x^{(i)}, y_w^{(i)}, y_l^{(i)})\}_{i=1}^N$.

To this day, \acrshort{rlhf} \citep{ouyang_training_2024} remains the most popular technique used in state-of-the-art \acrshort{llms} like GPT-4 \citep{openai_gpt-4_2023}, Claude \citep{bai_constitutional_2022}, Bard \citep{google_google_2023}, and Llama 2-Chat \citep{touvron_llama_2023}. Different implementations of \acrshort{rlhf} can vary in terms of data collection, training processes, and choice of \acrshort{rl} algorithms. Typically, \acrshort{rlhf} \citep{ouyang_training_2024} involves three main steps: (1) collecting feedback, (2) training a \acrshort{rm} based on that feedback, and (3) optimising the \acrshort{llms} using \acrshort{rl} techniques, such as \acrfull{ppo} \cite{schulman_proximal_2017}. Since \acrshort{rlhf} was first introduced, several advancements have been made, including fine-grained reward systems \citep{bai_constitutional_2022, wu_fine-grained_2023, dong_steerlm_2023, wang_helpsteer_2023, wang_helpsteer2_2024}, or replaced the original \acrshort{ppo} algorithm with other \acrshort{rl} techniques \citep{wu_pairwise_2023}.

An alternative to \acrshort{rlhf} is \acrshort{dpo} \citep{rafailov_direct_2023}, which employs an offline \acrshort{rl} approach to optimize language models based on preference data, without the need for a separate \acrshort{rm}. While DPO can be used independently, it is often complementary to other training methods like \acrshort{sft} or statistical rejection sampling, to further improve human alignment based on a \acrshort{rm} \citep{zhao_slic-hf_2023, liu_statistical_2024, dubey_llama_2024}. Statistical rejection sampling, also called best-of-N or top-k-over-N \citep{bai_constitutional_2022, touvron_llama_2023, dubey_llama_2024} is another widely used technique. Moreover, certain methods perform human alignment without \acrshort{rl} to avoid instabilities, and fine-tune the model on filtered samples by a \acrshort{rm}, or other sources \citep{dong_raft_2023, yuan_rrhf_2023}.

A major challenge in 
human alignment techniques is data acquisition \citep{casper_open_2023}. This includes issues 
such as evaluator misalignment, supervision difficulties, and feedback quality 
\citep{casper_open_2023}. However, as AI systems continue to improve, \acrshort{llms} are increasingly employed for tasks 
traditionally handled by humans, such as data annotation and generation. Unlike human feedback, AI-generated feedback offers better scalability, enabling faster and more cost-effective data collection. For example, \acrshort{rlaif}, introduced by \cite{bai_constitutional_2022}, is a promising approach that trains reward models based on preferences generated by off-the-shelf \acrshort{llms}. Variations of \acrshort{rlaif} have been explored in several studies \citep{lee_rlaif_2023, jiao_starling-7b_2023, cui_ultrafeedback_2024, li_hrlaif_2024, yang_rlcd_2024}. In the context of self-generating instructions, approaches like Self-Instruct \citep{wang_self-instruct_2023}, Self-Refine \citep{madaan_self-refine_2023}, and Self-Alignment \citep{li_self-alignment_2023} demonstrate how models can autonomously generate datasets based on their learned human preferences.

Different alignment methods like \acrshort{rlhf} and \acrshort{rlaif} rely on the \acrshort{rm} to incorporate the human feedback. The \acrshort{rm} learns to predict human preference based on labeled examples, serving as a proxy for human judgment later. Therefore, the success of language model alignment relies heavily on the quality of the underlying reward model \citep{pace_west--n_2024}, which in turn dictates the behaviour of the resultant chatbot \citep{shen_trickle-down_2023}. Even in \acrshort{llm} inference, methods like best-of-N sampling use the \acrshort{rm} to evaluate model outputs \citep{cui_ultrafeedback_2024}. \acrshort{rm} has also become crucial for generating synthetic data for preference alignment. In recent \acrshort{rlaif} methods, reward modeling has expanded beyond its traditional role and is now used to generate artificial feedback.

\subsubsection{Reward modeling} \label{rm_formulation}

In the original implementation \citep{ouyang_training_2024}, the goal of \acrshort{rm} training is to train a classifier that predicts the probability of human preference $p^{*}$ between two completions (\autoref{eq:3}), modelled by a Bradley-Terry model \citep{bradley_rank_1952}. The typical setup involves showing two completions, with preferences being measured using win-loss-tie outcomes or a Likert scale to capture the strength of preference \citep{bai_training_2022}. The data is processed into prompt-chosen-rejected trios, where the chosen completion, $y_{w}$, is preferred over the rejected one, $y_l$, forming the basis for training \citep{ouyang_training_2024}.

\begin{equation}
    \label{eq:3}
    p^{*}(y_w \succ y_l \mid x) = \frac{\exp(r^{*}(x, y_w))}{\exp(r^{*}(x, y_w)) + \exp(r^{*}(x, y_l))}.
\end{equation}

\begin{table}[t!]
\renewcommand{\arraystretch}{0.6}
\caption{\acrfull{et} features computed per word.}
\centering
\begin{adjustbox}{width=0.9\textwidth}
\small
\begin{tabular}{ll}
\toprule
\textbf{Feature} & \textbf{Definition} \\ \midrule 
 First Fixation Duration (FFD) & Time spent on the initial fixation\\
 Go-Past Time (GPT) &  Cumulative fixation time before moving to the right\\
 Total Reading Time (TRT) & Overall time spent fixating on a word\\
 Number of Fixations (nFix)& Number of fixations on each word\\ 
Proportion of participants (fixProp) & Proportion of participants that fixated on the word \\\bottomrule 
\end{tabular}
\end{adjustbox}
\label{tab:et-features}
\end{table}

\subsection{Eye-Tracking}\label{subsec:eye}

\acrfull{et} systems monitor oculomotor behavior, such as eye movements and fixations, offering valuable insights into visual attention, information processing, and expands our understanding of reading and language comprehension. \citep{zhang_eye-tracking_2024}. 
Specifically, \acrshort{et} data often include fixations -- pauses in eye movement to focus on specific areas \citep{mathias_survey_2020}; saccades -- rapid movements between two points \citep{mcguire_sentiment_2021}; scanpaths -- sequences of fixations that reveal saccades and regressions \citep{yang_plm-as_2023}; and other temporal and spatial gaze behavior features \citep{zhang_eye-tracking_2024}. Incorporating \acrshort{et} data into \acrshort{nlp} tasks often involves the use of several features listed in \autoref{tab:et-features}.

While several publicly available datasets such as ZUCO \citep{hollenstein_zuco_2020}, ZUCO2 \citep{hollenstein_zuco_2018}, PROVO \citep{luke_provo_2018}, ETSA-I \citep{mishra_predicting_2016}, ETSA-II \citep{mishra_cognition-cognizant_2018}, GECO \citep{cop_presenting_2017}, GECO-MT \citep{colman_geco-mt_2022} are widely used in \acrshort{et} research, obtaining real \acrshort{et} data for \acrshort{nlp} tasks remains a challenge. This is primarily due to the cost and precision requirements of \acrshort{et} equipment, the unavailability of gaze data during inference, as well as privacy concerns \citep{khurana_synthesizing_2023}. To address these challenges, two main approaches have been proposed. The first involves integrating \acrshort{et} data into the model during training through methods like \acrfull{mtl}, which eliminates the need for \acrshort{et} data during inference \citep{mishra_cognition-cognizant_2018, klerke_improving_2016, ren_cogalign_2021, yu_ceer_2024, deng_fine-tuning_2024}. The second approach involves techniques that directly predict users' gaze behaviour \citep{deng_fine-tuning_2024, deng_pre-trained_2023, zhang_eye-tracking_2024, wang_gaze-infused_2024}, creating synthetic \acrshort{et} data for any given text or stimulus \citep{deng_eyettention_2023,bolliger_scandl_2023,khurana_synthesizing_2023,li_torontocl_2021, hollenstein_cmcl_2021, hollenstein_cmcl_2022}.

\section{Related Work}

\textbf{Reward Modelling.}
The most popular approach to reward modeling follows the framework introduced by \cite{ouyang_training_2024}. Several studies have examined alternative versions for refining \acrshort{rm}s. For instance, \cite{bai_constitutional_2022} proposed more fine-grained reward structures, evaluating helpfulness and harmlessness separately. Other approaches have explored different reward modelling strategies \citep{wu_fine-grained_2023, dong_steerlm_2023, wang_helpsteer_2023}. Another line of research has focused on \acrfull{prms} \citep{lightman_lets_2024, uesato_solving_2022}  which differ from conventional \acrshort{rm}s by predicting the correctness of intermediate steps, rather than solely evaluating final outputs. Other studies implement data augmentation techniques \citep{shen_trickle-down_2023}, or cross-attention mechanisms between encoded input text and candidate pairs \citep{jiang_llm-blender_2023}. Moreover, some works have leveraged synthetic preference data for reward modelling \citep{cui_ultrafeedback_2024, jiao_starling-7b_2023}. \cite{wu_meta-rewarding_2024} built upon the LLM-as-a-Judge framework \cite{zheng_judging_2023} by introducing LLM-as-a-Meta-Judge, which evaluates the model’s judgments to generate preference pairs that enhance its decision-making capabilities. Finally, \cite{pace_west--n_2024} incorporated a self-training approach to improve reward model training. However, to date, no research has explored the integration of \acrshort{et} or other implicit feedback signals into \acrshort{rm}.

\textbf{\acrlong{et} in \acrlong{nlp}.} 
Several studies have investigated the use of \acrshort{et} data for a variety of \acrshort{nlp} tasks, such as named entity recognition \citep{hollenstein_entity_2019, ren_cogalign_2021, yu_ceer_2024, hollenstein_advancing_2019}, text comprehension \citep{ahn_towards_2020, reich_inferring_2022, sood_improving_2020}, language modelling \citep{huang_longer_2023, huang_long-range_2023, deng_eyettention_2023}, and question answering \citep{zhang_eye-tracking_2024, wang_gaze-infused_2024}. Other applications include code comprehension \citep{alakmeh_predicting_2024}, code summarization \citep{zhang_eyetrans_2024} and hallucination detection \citep{maharaj_eyes_2023}. \acrlong{et} has also been applied to sentiment analysis and sarcasm detection tasks \citep{mishra_predicting_2016, mishra_leveraging_2016, mishra_cognition-cognizant_2018, barrett_sequence_2018, huang_longer_2023, khurana_synthesizing_2023, hollenstein_advancing_2019, yang_plm-as_2023,kiegeland_pupil_2024, deng_pre-trained_2023, mathias_eyes_2018, mcguire_sentiment_2021}. The most relevant work to our approach is by 
\cite{kiegeland_pupil_2024}, which introduced a dataset generation method using \acrshort{et} signals for \acrshort{dpo}, building on the controlled sentiment generation framework proposed by \cite{deng_pre-trained_2023, yang_plm-as_2023}. While this study has contributed to the first steps towards integrating \acrshort{et} for human alignment in \acrshort{llms}, it is task- and dataset-specific, often relying on ranking criteria that underutilize the potential of \acrshort{et} feedback. In contrast, our approach presents a more general framework by directly incorporating implicit feedback into the \acrshort{rm}, rather than limiting its application to dataset creation.

\section{GazeReward: Reward modeling with \acrshort{et} Feedback} \label{sec:method}
\begin{figure}[htbp!]
    \begin{center}
    \includegraphics[width=\linewidth]{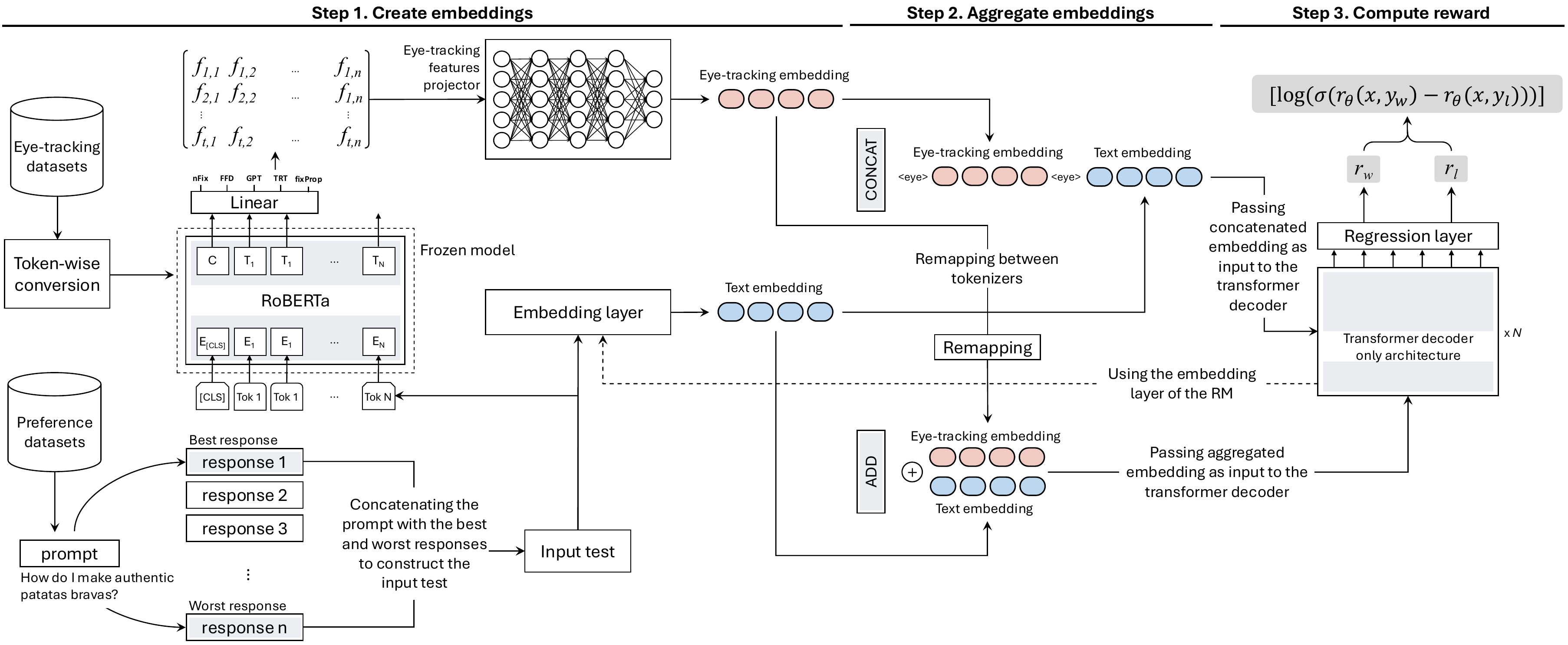}
    \end{center}
    \caption{Overview of the \textbf{GazeReward} framework, incorporating eye-tracking features into the reward model. The architecture is illustrated in the figure using the second \acrshort{et} prediction model, but it would be identical if the first one were used instead (see \autoref{sec:exp})}
    \label{fig:fig2}
\end{figure}

In this section, we discuss the proposed framework for augmenting the \acrshort{rm} using implicit feedback derived from \acrshort{et} signals (\autoref{fig:fig2}). Initially, we generate the \acrshort{et} features (\autoref{sec:gen_eye}) considering two state-of-the-art \acrshort{et} prediction models. Next, we combine the \acrshort{et} features with the text (\autoref{sec:add_eye}), producing different types of combined embeddings, and finally pass them as input into the \acrshort{rm} to obtain the reward for the prompt and its corresponding response (\autoref{sec:rm}).


\subsection{Eye-tracking features generation} \label{sec:gen_eye}
As discussed in \autoref{subsec:eye}, obtaining organic \acrshort{et} features for NLP applications presents several challenges. In this work, we consider an approach inspired by \acrshort{rlaif} research, where feedback is artificially generated from pre-trained LLMs and, in particular, from \acrshort{et} prediction models. Specifically, we incorporate the output of two different \acrshort{et} prediction models \citep{li_torontocl_2021,huang_long-range_2023} and evaluate the impact of different set of features. As input to these models, we pass the same text as we do in the \acrshort{rm}: a combination of prompt $x$ and response $y$. The output is a set of \acrshort{et} features, denoted as $f_{et}$, for each token  $f_{et} = \{f_1, f_2, \dots, f_w\} \in\mathbb{R}^{w \times f}$, where $w$ represents the number of tokens in the tokenizer used by the \acrshort{et} prediction model, and $f$ is the number of features. Depending on the specific model, between one and five synthetic features $f_{et} = \{f_1, f_2, \dots, f_w\} \in \mathbb{R}^{w \times f}$ are generated per token for the input text. 

\subsection{\acrshort{rm} augmentation using eye-tracking features}\label{sec:add_eye}
We implement two different approaches for incorporating \acrshort{et} features into the \acrshort{rm}, as shown in \autoref{fig:fig2}. In the first approach, \textbf{GazeConcat}, we concatenate the \acrshort{et} embeddings with the text embeddings. In the second approach, \textbf{GazeAdd}, we add the \acrshort{et} embeddings to the text embeddings. Furthermore, we concatenate the prompt and the response to be evaluated and pass them through the pre-trained embedding layer of \acrshort{rm}, to generate the embeddings $H = \{h_1, h_2, \dots, h_n\} \in \mathbb{R}^{n \times d}$, where $n$ is the number of tokens in the tokenizer used by the \acrshort{rm} and $d$ is the model embedding size. 

To project these features to the model embedding size ($d$), we use a \acrfull{mlp} \acrshort{et} feature projector $fp()$. The $fp()$ consists of two linear layers, two dropout layers, two Layer Normalization layers, and \texttt{ReLU} activation, designed for stable, non-linear \acrshort{et} feature representation and overfitting prevention. The model's input dimension dynamically adjusts to accommodate the number of features used during training. The \acrshort{et} features projector can be formulated as $emb_{ETF} = fp(f_{et}) \in \mathbb{R}^{w \times d}$ (Figure \ref{fig:fig2}). This formula describes the projection of ETF features ($f_{et}$) through a function $fp$, resulting in an embedding matrix $emb_{ETF}$ with dimensions $w \times d$, where $w$ represents the number of tokens and $d$ the embedding dimension.

\textbf{GazeConcat:}
The \acrshort{et} embedding, denoted as $emb_{ETF}$, is concatenated with the text embedding $H$ to form the input for the \acrshort{rm}. To distinguish between the two modalities, we introduce two special tokens: \textit{⟨eye⟩} and \textit{⟨/eye⟩}, which flag the start and end of the \acrshort{et} embedding, respectively (Figure \ref{fig:fig2}). These special tokens are randomly initialized as one-dimensional vectors and added to the embedding layer or the \acrshort{rm} model for training. The final input is structured as: $(emb(⟨eye⟩) \circ emb_{ETF} \circ emb(⟨/eye⟩) \circ H )$. The same process is applied to the attentions masks.

\textbf{GazeAdd:}
The input to the \acrshort{rm} consists of the \acrshort{et} embedding $emb_{ETF}$ and the text embedding $H$, which are added in an elementwise fashion:  $(emb_{ETF} + H)$. The two \acrshort{et} prediction models use different tokenizers, which also differ from those used by the base models in the \acrshort{rm}. As a result, the number of tokens $n$ in the input for the \acrshort{rm} and the number of tokens $w$ generated by the \acrshort{et} prediction model may not match. To address this embedding alignment issue, and have the same dimension, we remap the \acrshort{et} features from the $w$-token space to the $n$-token space used by each base model in the \acrshort{rm}. Further implementation details can be found in Appendix \ref{sec:remapping_appen}. 

\begin{table}[htbp!]
\renewcommand{\arraystretch}{0.4}
\caption{Overview of different corpora used in the study to train the reward model.}
\label{datasets_table}
\begin{center}
\begin{tabular}{lccccc}
\toprule
\multicolumn{1}{c}{\bf Corpus}  &\multicolumn{1}{c}{\bf Train set} &\multicolumn{1}{c}{\bf Val. set} &\multicolumn{1}{c}{\bf Test set} &\multicolumn{1}{c}{\bf Lang.} &\multicolumn{1}{c}{\bf Reference}
\\ \midrule
OASST1 & 6567 & 1160 & 416 &  EN$^*$ &  \cite{kopf_openassistant_2023}\\
HelpSteer2 & 5938 & 1049 & 364 &  EN & \cite{wang_helpsteer2_2024}\\\bottomrule 
\end{tabular}
\end{center}
\end{table}

\subsection{\acrlong{rm}} \label{sec:rm}


The \acrshort{rm}'s architecture and hyperparameters are identical to those of the pretrained \acrshort{llm}, except that the classification head used for next-token prediction is replaced with a regression head that outputs a scalar reward \citep{touvron_llama_2023}. This scalar reward indicates the quality of the model generation, corresponding to the predicted score for the final reply in a conversation. Differences in these rewards represent the log-odds that one response is preferred over another. The loss function is defined in \autoref{eq_rm}, where $y_{w}$ refers to the preferred response in a pair of completions $y_{w}$ and $y_l$. The dataset $D$ consists of human comparisons, where $r_\theta\left(x, y_w\right), r_\theta\left(x, y_l\right)$ represents the \acrshort{rm} $\theta$ scalar outputs for the preferred and less preferred completions, respectively \cite{ouyang_training_2024}.

\begin{equation}
\operatorname{loss}(\theta)=-E_{\left(x, y_w, y_l\right) \sim D}\left[\log \left(\sigma\left(r_\theta\left(x, y_w\right)-r_\theta\left(x, y_l\right)\right)\right)\right]
\label{eq_rm}
\end{equation}

In the proposed method, we augment the traditional \acrshort{rm}, which uses text input (a combination of the prompt $x$ and response $y$), by incorporating (artificial) implicit feedback through \acrshort{et} features generated from the same text. These \acrshort{et} features provide valuable information for capturing human preferences, thereby improving the \acrshort{rm}'s performance.

\section{Experiments}

\subsection{Experimental setup} \label{sec:exp}

\textbf{Datasets.}
For our experiments, we use the OpenAssistant Conversations dataset’s (OASST1) \citep{kopf_openassistant_2023} and HelpSteer2 \citep{wang_helpsteer2_2024} ( \autoref{datasets_table}). OASST1 is a human-generated, human-annotated, assistant-style conversation, created through global crowdsourcing and widely used for human alignment tasks \citep{kopf_openassistant_2023, dettmers_qlora_2023, wu_meta-rewarding_2024}. We filtered all non-English text, as the \acrshort{et} prediction models were exclusively trained on English data. Among the different responses in the dataset, we selected the two most distinct responses to compare the chosen and the rejected responses \citep{wang_self-taught_2024}. HelpSteer2 is a more recent, English-only dataset that has been used in studies such as \cite{wang_self-taught_2024, wang_helpsteer2_2024}. The dataset provides annotations for five response attributes: helpfulness, correctness, coherence, complexity, and verbosity. To transform it into a preference dataset, we designate the response with the higher helpfulness score as the chosen response and the other as the rejected response, following a method similar to that used in \acrshort{dpo} training \citep{wang_helpsteer2_2024} (see Appendix \ref{applychat} for more details about the datasets).

\textbf{Dataset Preparation.}
To tune LLMs for human-AI interaction, we need to define a chat dialogue protocol that allows the model to understand human instructions and rate them. To this end, we adopt a chat protocol that utilizes special header and termination tokens, similar to the format used in Llama 3. For example, in the case of the Llama 3 8B model, the concatenation of a prompt and its corresponding response would follow this structure: \textit{\textless im\_start\textgreater user} \textcolor{blue}{Example Prompt} \textit{\textless im\_end\textgreater} \textit{\textless im\_start\textgreater assistant} \textcolor{blue}{Example Response} \textit{\textless im\_end\textgreater} (see Appendix \ref{applychat} for more details).

\textbf{Models.}
 As \acrshort{rm} base models we use the pretrained checkpoint of Hugging Face (Appendix \ref{sec:app_models}) for Llama 3 8B, Llama 3 8B-instruct \citep{dubey_llama_2024} and Mistral 7B \citep{jiang_mistral_2023}.

\textbf{\acrshort{et} prediction models.}
In our analyses, we utilise two state-of-the-art \acrshort{et} prediction models, both pre-trained to predict \acrshort{et} features and kept frozen in our implementation. The input to these models is the same text used for the \acrshort{rm}, with minimal modifications (Appendix \ref{ref_predictor_appen}). The first model \citep{huang_long-range_2023}, consists of a T5 embedding layer \citep{raffel_exploring_2020}, a two-layer \acrshort{bilstm} \citep{hochreiter_long_1997}, and a one-hidden-layer \acrshort{mlp}. This model was trained on the Dundee, GECO \citep{cop_presenting_2017}, ZuCo1 \citep{hollenstein_zuco_2018}, and ZuCo2 \citep{hollenstein_towards_2020} datasets, and predicts total reading time (TRT) per token (\autoref{fig:trt_example}). The second model \citep{li_torontocl_2021}, is based on RoBERTa \citep{liu_roberta_2019} with a regression head on each token. This head is a linear layer that outputs five features: FFD, fixProp, GPT, TRT, and nFix (\autoref{tab:et-features}). The model is initialized with pre-trained weights and fine-tuned on the ZuCo1 \citep{hollenstein_zuco_2018}, ZuCo2 \citep{hollenstein_towards_2020} and PROVO \citep{luke_provo_2018} datasets. Since RoBERTa's maximum sequence length is 512 tokens and our input sequences are longer, we employ a sliding window approach. The input is split into 512-token segments with a 50-token overlap, and the results are combined using a linear weighted approach. Further details on these models and their integration into our framework are provided in Appendix \ref{ref_predictor_appen}.

\textbf{Baseline models.}
To evaluate the improvement in accuracy for a \acrshort{rm} that incorporates implicit feedback, and specifically \acrshort{et} signals, we compare the same \acrshort{rm} with and without the \acrshort{et} embeddings. For each dataset and model, we train and evaluate all combinations of integrating and combining \acrshort{et} features, and then compare them against a \acrshort{rm} trained on the same base model and dataset but without implicit feedback.

\textbf{Evaluation metrics.}
Performance is determined by measuring the model's ability to predict the better reponse from pairs of replies with different ranks. Accuracy is calculated as the percentage of cases where the reward score for the preferred response is higher than that of the less preferred response, based on a held-out dataset. This method follows similar approaches found in \cite{touvron_llama_2023, yuan_rrhf_2023, kopf_openassistant_2023, cui_ultrafeedback_2024}. We use the test split proposed by the authors for each dataset (\autoref{datasets_table}). We also conduct a complementary evaluation on RewardBench \citep{lambert_rewardbench_2024}, a benchmark dataset created for evaluating performance and safety features of \acrshort{rm}'s (see Appendix \ref{appen:rewardbench} for more details).


\textbf{Training procedure.}
In our implementetion the \acrshort{et} modules remains frozen (\autoref{fig:fig2}). For the \acrshort{rm}, we fine-tune the open-source models previously introduced with QLoRA \citep{dettmers_qlora_2023} a \acrfull{peft} method based on \acrfull{lora} \citep{hu_lora_2021}, with other memory optimization techniques. We follow the training process for the \acrshort{rm} as outlined in \cite{touvron_llama_2023, ouyang_training_2024}. We independently train each model on its respective dataset for two epochs, as detailed in \cite{wang_helpsteer2_2024}. For hyperparameter tuning, we reserve 15\% of each dataset for validation (shown in \autoref{datasets_table}). The best-performing checkpoints are selected based on the lowest validation loss and used for performance evaluation. We perform a grid search to determine the optimal batch size and testing values of \{8, 16, 32\}. The AdamW optimizer \citep{loshchilov_decoupled_2019} is used, with the \acrfull{lr} is tuned over the range \{1, 5, 10, 50\}e-6, following the values reported in \cite{touvron_llama_2023, cui_ultrafeedback_2024, wang_helpsteer2_2024}. Additionally, we evaluate different \acrshort{lr} schedulers: constant, linear, and cosine with a minimum LR. Further hyperparameter values and implementation details for both the \acrshort{rm} and the \acrshort{et} projector can be found in Appendix \ref{sec:hypertuning}.

\subsection{Results} \label{sec:results}

The results of our experiments on the OASST and HelpSteer datasets, covering all possible combinations of \acrshort{et} features, models, and inclusion methods, as shown in \autoref{main_oasst} and \autoref{main_helpsteer} respectively. For the Mistral model, results for the \textbf{GazeAdd} method are unavailable due to the inability to map features between the \acrshort{et} prediction model's tokenizer and the reward model's tokenizer (details in Appendix \ref{sec:remapping_appen}). In what follows, we present key findings based on three seeds, reporting the average, mean, and statistical significance.

\begin{table}[htbp!]
\renewcommand{\arraystretch}{0.4}
\caption{Reward modeling accuracy (\%) for OASST1 dataset. The highest results are in bold and the second highest are underlined.}
\label{main_oasst}
\begin{center}
\resizebox{\textwidth}{!}{%
\begin{tabular}{cccccccc}
\toprule
   & & \multicolumn{2}{c}{\bf Llama-3-8B-Instruct} & \multicolumn{2}{c}{\bf Llama-3-8B} & \multicolumn{2}{c}{\bf Mistral-7B} \\
\midrule
 & baseline &65.9 $\pm$ 0.5${}$ & diff (\%) & 65.5 $\pm$ 2.1${}$ & diff (\%) & 66.3 $\pm$ 0.1${}$ & diff (\%) \\
\midrule
\multirow{3}{*}{\textbf{GazeConcat}}
& $fcomb_{1}$ & 69.0 $\pm$ 0.4${*}$ & 4.7 & 69.3 $\pm$ 0.6${}$ & 5.9 & 67.6 $\pm$ 1.7${}$ & 2.1 \\
& $fcomb_{2.5}$ & \underline{70.2 $\pm$ 0.3${**}$} & 6.6 & \textbf{71.5 $\pm$ 0.5${}$} & 9.2 & \underline{70.2 $\pm$ 0.4${*}$} & 5.9 \\
& $fcomb_{2.2}$ & 70.0 $\pm$ 0.4${**}$ & 6.3 & \underline{71.2 $\pm$ 0.8${}$} & 8.8 & \textbf{71.0 $\pm$ 1.0${}$} & 7.1 \\
\midrule
\multirow{3}{*}{\textbf{GazeAdd}}
& $fcomb_{1}$ & 68.9 $\pm$ 0.9${}$ & 4.6 & 68.9 $\pm$ 1.0${}$ & 5.3 & -  & \\
& $fcomb_{2.5}$ & \textbf{70.2 $\pm$ 0.1${*}$} & 6.6 & 69.5 $\pm$ 0.3${}$ & 6.1 & -  & \\
& $fcomb_{2.2}$ & 69.0 $\pm$ 0.4${*}$ & 4.7 & 68.3 $\pm$ 0.7${}$ & 4.4 & -  & \\
\bottomrule
\end{tabular}}
\end{center}
\end{table}

\begin{table}[htbp!]
\renewcommand{\arraystretch}{0.4}
\caption{Reward modeling accuracy (\%) for Helpsteer2 dataset. The highest results are in bold and the second highest are underlined.}
\label{main_helpsteer}
\begin{center}
\resizebox{\textwidth}{!}{%
\begin{tabular}{cccccccc}
\toprule
   & & \multicolumn{2}{c}{\bf Llama-3-8B-Instruct} & \multicolumn{2}{c}{\bf Llama-3-8B} & \multicolumn{2}{c}{\bf Mistral-7B} \\
\midrule
 & baseline &54.7 $\pm$ 0.7${}$ & diff (\%) & 53.3 $\pm$ 0.8${}$ & diff (\%) & 54.1 $\pm$ 0.3${}$ & diff (\%) \\
\midrule
\multirow{3}{*}{\textbf{GazeConcat}}
& $fcomb_{1}$ & \underline{61.1 $\pm$ 1.2${*}$} & 11.8 & 59.1 $\pm$ 0.2${*}$ & 10.8 & \underline{57.6 $\pm$ 2.3${}$} & 6.5 \\
& $fcomb_{2.5}$ & 58.5 $\pm$ 1.6${}$ & 7.0 & \underline{60.3 $\pm$ 0.5${**}$} & 13.2 & \textbf{58.7 $\pm$ 2.4${}$} & 8.5 \\
& $fcomb_{2.2}$ & 60.6 $\pm$ 3.3${}$ & 10.9 & 57.9 $\pm$ 2.0${}$ & 8.6 & 56.0 $\pm$ 2.4${}$ & 3.4 \\
\midrule
\multirow{3}{*}{\textbf{GazeAdd}}
& $fcomb_{1}$ & \textbf{62.3 $\pm$ 0.6${**}$} & 13.9 & \textbf{62.4 $\pm$ 1.0${**}$} & 17.0 & - &  \\
& $fcomb_{2.5}$ & 59.6 $\pm$ 1.1${*}$ & 9.0 & 58.6 $\pm$ 1.2${*}$ & 10.0 & - &  \\
& $fcomb_{2.2}$ & 60.3 $\pm$ 0.5${**}$ & 10.2 & 59.3 $\pm$ 0.1${*}$ & 11.3 & - &  \\ \bottomrule
\end{tabular}}
\end{center}
\end{table}

\textbf{Effect of Model Initialization.}
We evaluate the impact of model initialization on performance. Open-access \acrshort{llms} typically come in two forms: a pre-trained version without human alignment and a final version that has undergone alignment with human feedback. Since we lack access to intermediate checkpoints, we experiment with both pre-trained models (Mistral 7B and Llama 3) and models that are already human-aligned (Llama 3 Instruct). Our goal is to confirm that our method is effective for \acrshort{rm} initialized with both pre-trained and human-aligned checkpoints. When comparing accuracy improvements relative to the baseline (without \acrshort{et} features), all models show considerable gains from incorporating implicit feedback. Notably, the Llama 3 8B and Mistral 7B models, which had no prior alignment, demonstrate performance improvement from the incorporation of \acrshort{et} features, indicating that unaligned models can benefit from implicit feedback.

\textbf{Inclusion method.}
The results shown \autoref{main_oasst} and \autoref{main_helpsteer} indicate that both \textbf{GazeConcat} methods and \textbf{GazeAdd} introduce a substantial performance improvements to the \acrshort{rm}. Across both datasets, concatenating embeddings (\textbf{GazeConcat}) delivers more consistent results. Incorporating \acrshort{et} information through specialized separator embeddings allows the model to process both text and \acrshort{et} features more robustly. However, in the HelpSteer dataset (\autoref{main_helpsteer}), directly adding \acrshort{et} information to the text embeddings (\textbf{GazeAdd}) results in the greatest improvement over the baseline.

\textbf{\acrfull{et} feature importance.}
Different \acrshort{et} features capture distinct aspects of reading behaviour and information processing, influencing model performance uniquely \citep{zhang_eye-tracking_2024}. Here, we examine how model performance varies when incorporating three different feature combinations generated by two different \acrshort{et} prediction models: $fcomb_{1}$ -- TRT generated by the first \acrshort{et} prediction model; $fcomb_{2.5}$ -- five features (nFix, FFD, GPT, TRT, fixProp) generated by the second \acrshort{et} prediction model; and $fcomb_{2.2}$ --  TRT and FFD generated by the second \acrshort{et} prediction model. TRT and FFD are widely used in \acrshort{et} research \citep{huang_longer_2023, huang_long-range_2023, zhang_eye-tracking_2024, maharaj_eyes_2023, wang_semgraph_2022}, and they have been shown to correlate with attention scores from pre-trained transformer models \citep{wang_gaze-infused_2024, bensemann_eye_2022, sood_interpreting_2020} and with gradient-based saliency\citep{hollenstein_relative_2021, wu_eye_2024}. When comparing results, we observe that the \acrshort{rm} benefits from implicit feedback regardless of the \acrshort{et} feature combination or \acrshort{et} prediction model used. Specifically, in most cases, $fcomb_{1}$ yields the best results, particularly with the \textbf{GazeAdd} method. For \textbf{GazeConcat}, $fcomb_{2.2}$ and $fcomb_{2.5}$ performs best in general. We attribute the superior performance of $fcomb_{1}$ to how the \acrshort{et} prediction model generating the fixations was trained, including the data and preprocessing methods used (see Appendix \ref{ref_predictor_appen}). Moreover, when comparing $fcomb_{2.2}$ and $fcomb_{2.5}$ -- both generated by the same model -- only in one case does integrating nFix, GPT, and fixProp improves performance. In some instances, using $fcomb_{2.5}$ results in worse performance than the baseline, confirming findings provided by prior studies, which suggest that features related to reading time, such as FFD and TRT, contribute most to performance gains.



\textbf{RewardBench.}
As a side contribution, we evaluate our best performing models (trained on the OAAST1 dataset) on RewardBench. This evaluation is not intended to directly compare our method with larger, more resource-intensive \acrshort{rm}, but rather to show that through the integration of multimodal signals like \acrshort{et} features we can significantly enhance the performance of \acrshort{rm} models. 
The results shown in \autoref{table_results_reward} demonstrate consistent improvements as previously observed, with gains exceeding 40\% for the Mistral model -- a notable gain considering that the base RM is the same. 
We note that the performance of the baseline models is impacted by \acrshort{rm} trained on base models with less than 9B parameters and on relatively small datasets (see details in Appendix \ref{appen:rewardbench}).




\begin{table}[htbp!]
\renewcommand{\arraystretch}{0.4}
\caption{Reward modeling accuracy (\%) evaluating on RewardBench dataset. All models are trained on OASST1 dataset. The highest results are in bold and the second highest are underlined.}
\label{table_results_reward}
\begin{center}
\begin{tabular}{cccccccc}
\toprule
   & & \multicolumn{2}{c}{\bf Llama-3-8B-Instruct}  & \multicolumn{2}{c}{\bf Llama-3-8B} & \multicolumn{2}{c}{\bf Mistral-7B} \\ \midrule 
& baseline & 46.9  & diff(\%)  & 50.9  &  diff(\%)  & 41.2 & diff(\%) \\  \midrule 
\multirow{3}{*}{\textbf{GazeConcat}}
&  $fcomb_{1}$  & 57.8  & 23.1\%  & \underline{58.4}  & \underline{14.5\%}  & 59.9  & 45.4\% \\ 
& $fcomb_{2.5}$ & \textbf{58.4}  & \textbf{24.4\%}  & 58.1  & 14.1\%  & \underline{60.3}  & \underline{46.4\%} \\ 
& $fcomb_{2.2}$ & \underline{58.1}  & \underline{23.8\%}  & \textbf{58.5}  & \textbf{14.8\%}  & \textbf{60.5}  & \textbf{46.9\%} \\ \midrule
\multirow{3}{*}{\textbf{GazeAdd}} 
& $fcomb_{1}$   & 56.5  & 20.3\%  & 56.6  & 11.2\%  & -     & \\ 
& $fcomb_{2.5}$ & 54.9  & 16.9\%  & 53.8  & 5.6\%  & -     & \\ 
& $fcomb_{2.2}$ & 55.4  & 17.9\%  & 52.5  & 3.1\%     & -     &  \\ \bottomrule
\end{tabular}
\end{center}
\end{table}

\section{Discussion} \label{sec_discussion}

In this work, we introduced a novel framework for integrating implicit feedback into the \acrlong{rm}, a key component for aligning \acrshort{llms} and generating synthetic data for further alignment. We validated our approach using widely-adopted, open-source models such as Llama 3 and Mistral, for initializing the \acrshort{rm}. By employing two different models to generate \acrshort{et} features, our results show that incorporating implicit feedback consistently improves the \acrshort{rm}'s ability to predict user preferences, regardless of the model used and without the need to reach large parameter counts or train on massive datasets. Additionally, our method leverages \acrshort{et} features generated by models, making it fully scalable and applicable to various human alignment methods, including those that involve artificially generated datasets. This work advances the ongoing discussion on optimizing AI alignment with human values and shows the potential of multimodal signals for \acrshort{nlp} research.

\subsection{Limitations \& Future Work}

\textbf{Data}. A limitation of our study is that both \acrshort{et} prediction models were trained on a relatively small datasets (Appendix \ref{ref_predictor_appen}) that are not tailored to our tasks. Future work could benefit from directly collecting \acrshort{et} data specifically for \acrshort{llm}-generated responses, to offer insights into human reading comprehension and information processing of prompts, which could further improve model performance. Additionally, since the \acrshort{et} prediction models used in our experiments were trained on English corpora, the method’s generalizability to other languages requires further investigation. Moreover, we explored two methodologies for integrating \acrshort{et} features into the \acrshort{rm}, but other approaches could prove more effective. For instance, \acrshort{et} features could be used to modify the \acrshort{rm}’s attention mask, as suggested by \cite{zhang_eye-tracking_2024}. Regarding dataset selection, both models used, Mistral 7B \citep{jiang_mistral_2023} and Llama 3 \citep{dubey_llama_2024}, were fine-tuned on publicly available data, though specific details on the datasets are limited. Therefore, we cannot discount the possibility that the datasets we used may have been part of the models’ pretraining, particularly for Llama 3 7B Instruct, which has already undergone human alignment. However, as we compare against baselines using the same model checkpoints, any potential effects would be consistent across both conditions. 

\textbf{Training}. 
The scaling trends for the \acrshort{rm} \citep{touvron_llama_2023} show that larger models or models trained on massive datasets perform better. A promising direction would be to evaluate our framework on larger models, without relying on \acrshort{peft} methods, and on larger datasets. However, this would incure significant computation costs. Another direction is integrating the proposed \acrshort{rm} into an alignment method like \acrshort{rlhf}, or applying it in rejection sampling to generate synthetic preference datasets, ensuring that accuracy gains in the \acrshort{rm} translate to improvements in the final \acrshort{llms}.

\subsubsection*{Reproducibility statement}
All the code necessary to reproduce is in the GitHub repository \footnote{\url{https://github.com/Telefonica-Scientific-Research/gaze_reward}}. Both datasets used are publicly available \citep{kopf_openassistant_2023, wang_helpsteer2_2024}. Additionally, both \acrshort{et} prediction models have been trained with public datasets \citep{cop_presenting_2017, hollenstein_zuco_2018, hollenstein_towards_2020}.

\subsubsection*{Impact Statement}
Since our research uses only synthetic \acrshort{et} data, there are no privacy concerns or need for large-scale experiments involving human subjects. We should also raise attention to the limitations of human feedback and ET prediction models bias, that inevitably reflect aspects of their training data.

\subsubsection*{Acknowledgments}
This research is supported by Horizon Europe's European Innovation Council through the Pathfinder program (SYMBIOTIK project, grant 101071147) and by the Industrial Doctorate Plan of the Department of Research and Universities of the Generalitat de Catalunya, under Grant AGAUR 2023 DI060. We also want to thank Sebastian Macaluso for his important feedback during the project.

\bibliographystyle{unsrtnat}
\bibliography{references}

\appendix
\section{Appendix}
\subsection{Implementation details}
This section provides further details on the implementation of our method. Subsection \ref{applychat} provides more details on the datasets used and their preprocessing steps. In subsection \ref{ref_predictor_appen}, further information is given about the models used for generating \acrshort{et} features, along with the specific preprocessing required for each. Subsection \ref{sec:remapping_appen} explains the process of mapping the fixations from the tokenizer used by the generation model to the tokenizer used by the Reward Model.  Subsection \ref{sec:app_models} give more detials on the checkpoints used fot the \acrshort{rm} backbone models. Finally, additional implementation details are discussed in subsection \ref{training}.

\subsubsection{Dataset processing}\label{applychat}

In this subsection, we provide more details about the datasets used and the preprocessing to train the \acrshort{rm}. We use two different datasets: OpenAssistant Conversations dataset’s (OASST1) \footnote{\url{https://huggingface.co/datasets/OpenAssistant/oasst1}}  \citep{kopf_openassistant_2023} and HelpSteer2 \footnote{\url{https://huggingface.co/datasets/nvidia/HelpSteer2}} \citep{wang_helpsteer2_2024}.

\textbf{OASST1}. A human-generated, human-annotated assistant-style conversation corpus consisting of 161,443 messages in 35 different languages, resulting in over 10,000 complete and fully annotated conversation trees. The basic data structure is a \acrfull{ct}, with nodes representing written messages in a conversation. A \acrshort{ct}’s root node represents an initial prompt, given by the prompter. The data was collected using a web-app interface as a product of a worldwide crowd-sourcing effort involving over 13,500 volunteers, dividing the collection of each tree into five separate steps: prompting, labelling prompts, adding reply messages as prompter or assistant, labelling replies, and ranking assistant replies.

\textbf{HelpSteer2.} A CC-BY-4.0-licensed open-source helpfulness dataset, designed to train state-of-the-art \acrshort{rm} consisting on 10,000 response pairs. It collects prompts mainly from ShareGPT \footnote{\url{https://huggingface.co/datasets/RyokoAI/ShareGPT52K}}, focusing on user inputs and filtering out non-English and programming-related prompts for quality. The prompts are clustered into topics and sampled based on complexity to ensure diversity. Multi-turn prompts are generated using an in-house model, with responses sourced from various internal models and human annotators. For each response, they annotate five attributes (helpfulness, correctness, coherence, complexity, and verbosity) on a Likert-5 scale involving multiple annotators for each response, ensuring high-quality ratings across five attributes.

\textbf{Conversation format and dataset preparation.}
\\
To fine-tune \acrshort{llms} for human-AI interaction, we need to define a chat protocol. We use a multi-message chat setup with a special header and termination tokens, similar to the one in Llama 3 \cite{dubey_llama_2024}. The header tokens differentiate the turns between the user and the system. For this, we use the \textit{apply\_chat\_template}\footnote{\url{https://huggingface.co/docs/transformers/main/en/chat_templating}} feature from \textit{FastTokenizers} in the \textit{transformers} library.  

The tokenizer used by the Meta-Llama-3-8B-Instruct model already incorporates this chat format since this model has already undergone human alignment. Therefore, we use this format in our experiments. For the other two models, we employ the default chat format provided by their respective tokenizers. We add new tokens in the embeddings layer for these chat formats and we train them as part of our process. Below, we provide an example of the template for each model.

\begin{itemize}

    \item \textbf{Meta-Llama-3-8B-Instruct:}
    \textit{\textless begin\_of\_text\textgreater \textless start\_header\_id\textgreater user\textless end\_header\_id\textgreater} \textcolor{blue}{Example Prompt} \textit{\textless eot\_id\textgreater \textless start\_header\_id\textgreater assistant\textless end\_header\_id\textgreater} \textcolor{blue}{Example Response} \textit{\textless eot\_id\textgreater}

    \item \textbf{Meta-Llama-3-8B:} \textit{\textless im\_start\textgreater user} \textcolor{blue}{Example Prompt} \textit{\textless im\_end\textgreater} \textit{\textless im\_start\textgreater assistant} \textcolor{blue}{Example Response} \textit{\textless im\_end\textgreater}

    \item \textbf{Mistral-7B:} \textit{\textless s\textgreater [INST]} \textcolor{blue}{Example Prompt} \textit{[/INST]} \textcolor{blue}{Example Response} \textit{\textless/s\textgreater}
\end{itemize}

\subsubsection{Eye-tracker features generation models} \label{ref_predictor_appen}
Special tokens are removed from the text before it is tokenized with the corresponding tokenizer used for the \acrshort{et} generator model. This is done to ensure that special tokens related to the chat format are not included in the input and are not assigned \acrshort{et} features to them, since these tokens are just for the \acrshort{rm} to understand the chat format.

\begin{figure}[tb!]
    \centering
    \begin{minipage}[b]{0.46\textwidth}
        \centering
        \includegraphics[width=\linewidth]{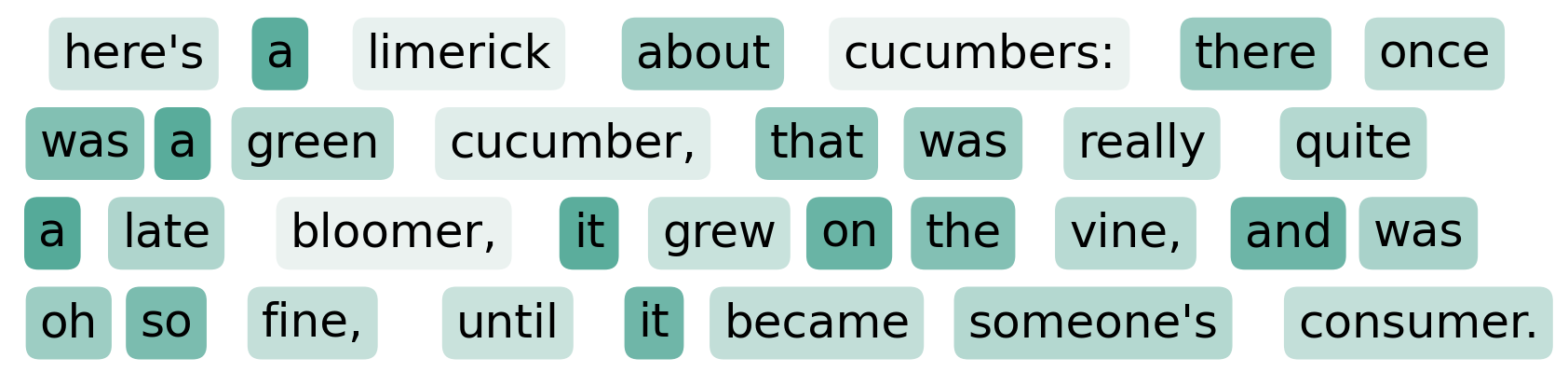}
        \caption*{TRT per word in chosen response.}
    \end{minipage}
    \begin{minipage}[b]{0.46\textwidth}
        \centering
        \includegraphics[width=\linewidth]{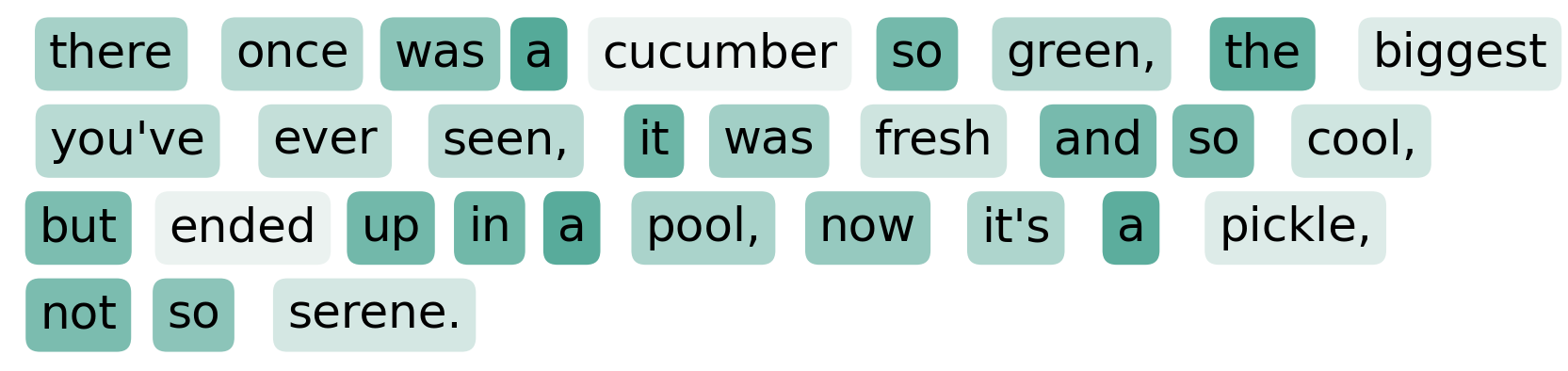}
        \caption*{TRT per word in rejected response.}
    \end{minipage}
    \caption{TRT generated by first model \citep{huang_long-range_2023} of the chosen and rejected response to prompt 'Create a limerick about cucumbers'. Deeper colour represents longer fixation.}
    \label{fig:trt_example}
\end{figure}

\textbf{First model:} Model presented in  \cite{huang_long-range_2023}. The code for the model along with the weights is publicly available, so we used the pre-trained checkpoint and we adapted their code for our implementation. This model was trained on several eye-tracking datasets, including Dundee \citep{kennedy_frequency_2012}, GECO \citep{cop_presenting_2017}, ZuCo1 \citep{hollenstein_zuco_2018}, ZuCo2 \citep{hollenstein_towards_2020}. More detailed information about this datasets is presented in \autoref{tab:corpus_et_overview}. The best model achieves an \acrfull{mse} of 4.02 on a randomly held-out test set (25\% of all data). This model has 17.5M pararemetes and remains frozen during training. In \autoref{fig:trt_example} we show an example of the synthetic \acrfull{trt} generated for the chosen and rejected response to a prompt. 

For training this model, since the fixation duration is distributed differently across corpora, the authors normalize the fixation duration for each corpus, by dividing it by the mean duration of the corpus. Moreover, they map the duration values to discrete space $[1, 2, · · · , K]$. Using K-quantiles, the fixation values were partitioned into $K$ subsets of nearly equal sizes, and each value was assigned to the index of the corresponding subset. The model is then trained in a multi-task setting, computing the mean and variance of the fixation duration. This quartile-based processing is used in other works \citep{huang_longer_2023} that use \acrshort{et} data to improve performance in \acrshort{nlp} tasks, and we believe it is part of the reason why we obtained better results with this combination when training the \acrshort{rm}. The authors proposed a method specifically for converting word-level TRT to token-level fixation data during model training. Initially, the TRT of a word is assigned to its characters, then a small number is assigned to the last character of the word (mainly to give small values to punctuation). After tokenizing the word the span of each subword is obtained, and the maximum value in each span is taken as the final token-level fixation data.

To use this model in our setup, we need to reverse this conversion process and recompute the features from token-level back to word-level, allowing us to remap the features to a different tokenizer. This is done by summing the orignal features for all tokens corresponding to a word and then distributing them across the tokens mapped to the same word in the other tokenizer. More details of this conversion are explained in Appendix \ref{sec:remapping_appen} and an example of the process in \autoref{tab:remapping}.

\textbf{Second model:} Model presented in \cite{li_torontocl_2021}. The code and training data are also publicly available, so we trained it following their original methodology and we adapted their code to incorporate it into our implementation. The model was trained using the  ZuCo 1 \citep{hollenstein_zuco_2018} and ZuCo 2 \citep{hollenstein_towards_2020} and PROVO \citep{luke_provo_2018} datasets. For the ZuCo datasets, 800 sentences (15.7 tokens) were provided as training data and 191 sentences (3.5k tokens) were held out for evaluation. Information about the datasets used to train these models is in \autoref{tab:corpus_et_overview}. The model is based on RoBERTa \citep{liu_roberta_2019} with a regression head on each token. This head is a linear layer that outputs five features: FFD, fixProp, GPT, TRT, and nFix (\autoref{tab:et-features}). \acrfull{mae} for each feature is presented in \autoref{tab:mae_model2}. This model has 125M parameters and remains frozen during training.

In this generative model, the conversion of word-level features to token-level features during training is done by assigning the features of a word to the first token and it is assumed that the rest of the tokens of this word do not have features. We reversed this process similarly during inference by forcing the predictions for tokens that are not the first in a word to be zero.  Since the maximum sequence length for RoBERTa is 512 and we are dealing with longer sequences, we implemented a sliding window approach. We split the input into sequences of 512 tokens with a 50-token overlap. After processing, we combine the results using a linear combination. 

\begin{table}[h]
    \centering   
    \caption{MAE performance of the model reported in \cite{li_torontocl_2021}.}
      \label{tab:mae_model2}
    \begin{tabular}{cccccc}
    \hline
     \textbf{nFix} & \textbf{FFD} & \textbf{GPT} & \textbf{TRT} & \textbf{fixProp} & \textbf{All (Dev)}  \\ \hline
    3.984 & 0.713 & 2.424 & 1.556 & 10.781 & 3.892 \\ \hline
    \end{tabular}
    
\end{table}

\begin{table}[h]
    \centering
    \caption{Overview of different corpora used to train the \acrshort{et} features generator models.}
    \label{tab:corpus_et_overview}
    \begin{center}
    \begin{tabular}{c|ccccc}
        \textbf{Corpus} & \textbf{Lang.} & \textbf{Sents.} & \textbf{Tokens} & \textbf{Subjects} & \textbf{Reference} \\ \hline \\
        Dundee     & EN &  2367  & 58598  & 20 & \cite{kennedy_frequency_2012} \\
        Provo      & EN & 189 & 2659 & 84 & \cite{luke_provo_2018} \\
        ZuCo 1     & EN & 300 & 6588 & 12 & \cite{hollenstein_zuco_2018} \\
        ZuCo 2     & EN & 349 & 6828 & 18 & \cite{hollenstein_zuco_2020} \\
        Geco       & EN* & 2449 & - & 23 & \cite{cop_presenting_2017}  \\ \hline 
    \end{tabular}
\end{center}
\end{table}

\subsubsection{Mapping \acrshort{et} features between different tokenizers} \label{sec:remapping_appen}

Both \acrshort{et} features generator models used are based on different tokenizers, which are also different from the tokenizers employed by the based models used as \acrshort{rm}. As a result, the number of tokens $n$ in the input for the reward model and the number of tokens $w$ for the \acrshort{et} features may not be the same. For \textbf{GazeAdd}, to be able to combine elementwise the \acrshort{et} feature embedding and the text embedding, they must have the same temporal dimensions. Therefore, we need to map the \acrshort{et} features per token from the \acrshort{et} tokenizer to the tokens of the \acrshort{rm} tokenizer. Specifically, we convert our $f_{et} \in \mathbb{R}^{w \times f}$ ($w$ is the number of tokens, and $f$ is the number of features) to $f_{et}^{mapped} \in \mathbb{R}^{n \times f}$  where $n$ is the number of tokens in the \acrshort{rm} input. For that. we perform a mapping between the two tokenizers to obtain the \textit{mapped features} $f_{et}^{mapped}$.

To map tokens generated by two different tokenizers, we use our \textit{EyeTrackPy} python library that will be publicly released. First, we perform an initial mapping of tokens to the words they belong to in each tokenizer with some properties of \textit{FastTokenizers} from the \textit{transformers} library \footnote{\url{https://huggingface.co/docs/transformers/main_classes/tokenizer}}. Then, we map words from one tokenizer to the words in the other and finally, we assume that the combination of the tokens that are mapped to a word in one tokenizer correspond to the tokens that are mapped to the word that is mapped to the initial word in the other tokenizer. Each row in \autoref{tab:remapping}, refers to a step in this process. 

For each predictor, we reverse the method used to convert word-level features into token-level features (more details in Appendix \ref{ref_predictor_appen}) but passing from tokens in the first one, to tokens in the second tokenizer. For example, if for the first \acrshort{et} features predictor models tokens $t_{1},t_{2}$ are mapped to tokens $t_{1},t_{2},t_{3}$ in another second tokenizer, the values sum for all the tokens in the first list and distribute them equally across all the tokens in the second list: being $t_{1}$ (1s TRT) and $t_{2}$ (2s TRT) each of $t_{1},t_{2},t_{3}$ are assigned a TRT of $(1+2)/3=1s$. In  \autoref{tab:remapping} is represented a example of this process where row TRT(1) are the final TRT mapped for the first \acrshort{et} predictor and TRT(2) for the second one. Finally, because special chat tokens were removed when generating the \acrshort{et} features, we assign value $0$ for all features in this tokens. At the time of publishing this work, some of the tokenizer functionalities needed for alignment between tokenizers were not available in Mistral 7B.

\begin{table}[h]
    \centering
    \caption{Example of mapping TRT between two different tokenizers. TRT (1) represents the process used for the first \acrshort{et} predictor, and TRT(2) for the second \acrshort{et} predictor.}
    \label{tab:remapping}
    \begin{center}
    \begin{tabular}{c|ccccc}
         & \textbf{Tokenizer 1} & \textbf{Tokenizer 2} \\ \hline
        Words & astrophotography & astrophotography \\
        Tokens str& ['\_Astro', 'photo', 'graphy'] &[$\dot{C}$, 'Ast', 'roph', 'ot', 'ography'] \\
        Tokens idx & \texttt{[22, 23, 24]} & \texttt{[23, 24, 25, 26, 271]} \\
        Tokens IDs & \texttt{[15001, 17720, 16369]} & \texttt{[198, 62152, 22761, 354, 5814]} \\
        TRT (1) & \texttt{[11.23, 11.49, 10.16]} & \texttt{[6.58, 6.58, 6.58, 6.58, 6.58]} \\
        TRT (2) & \texttt{[24.53, 0, 0]} & \texttt{[24.53, 0, 0, 0, 0]} \\
        \hline
    \end{tabular}
\end{center}
\end{table}

\subsubsection{Models}  \label{sec:app_models}
 As \acrshort{rm} base models we use the pretrained checkpoint of Hugging Face for Llama 3 8B \footnote{\url{https://huggingface.co/meta-llama/Meta-Llama-3-8B}} \citep{dubey_llama_2024}, Llama 3 8B-instruct \footnote{\url{https://huggingface.co/meta-llama/Meta-Llama-3-8B-Instruct}}\citep{dubey_llama_2024} and Mistral 7B \footnote{\url{https://huggingface.co/mistralai/Mistral-7B-v0.3}} \citep{jiang_mistral_2023}.

\subsubsection{Training datails} \label{training}

During training, the \acrshort{et} features predictor model remains frozen. The \acrshort{rm} is fine-tuned on top of the open-source models using QLoRA \citep{dettmers_qlora_2023} based on \acrfull{lora} \citep{hu_lora_2021}, which fine-tunes select dense layers by optimizing low-rank decomposition matrices representing weight changes, instead of directly updating pre-trained weights. QLoRA introduces memory optimization techniques such as the 4-bit NormalFloat (NF4), a novel data type, to improve performance without increasing memory usage.  Following \cite{dettmers_qlora_2023} we use hyperparameters: r=8, alpha=32, and dropout=0.1.

We also fine-tune the \acrshort{rm} embedding layer, since we are adding new tokens for the chat format and special separators tokens in our \textbf{RewardConcat} method (\autoref{sec:method}). Also, the last layer added to the \acrshort{rm} for the scalar reward is trained from scratch without adapters. Our implementation is based in \textit{pytorch} and we use \textit{transformers} \footnote{\url{https://huggingface.co/docs/transformers/index}} from Hugging Face. 

\textbf{Hardware.}
We trained the models on servers equipped with 2x Intel Xeon Platinum 8470 CPUs, 1TB of RAM, and either 2x NVIDIA H100 (80GB) or 4x NVIDIA A100 (80GB) GPUs. We always train using only GPU at a time per each model and training times were between 20 and 50 hours depending mainly on the number of steps between model evaluations.

\section{Reward benchmark}\label{appen:rewardbench}
As we described in \autoref{sec_discussion}, a future direction would be to train a Reward Model on a larger model with more data. It has been proven the scaling trends for the reward model; More data and a larger-size model generally improve accuracy \citep{touvron_llama_2023}. Nevertheless, as a complement to our results, we evaluated our trained models on the dataset with the best results, OAAST1, in this RewardBench, a benchmark for Reward Models.
RewardBench, proposed in \cite{lambert_rewardbench_2024}, is a benchmark designed to evaluate the performance and safety of reward models. It consists of a set of datasets intended for measuring how reward models perform on challenging prompts across chat, reasoning, and safety domains, using a trio structure of prompt-chosen-rejected pairs. It comprises 2985 diverse tasks, each sample is formatted as a prompt with a manual or human-verified chosen and rejected completion. Due to its diversity of tasks (4 categories and 23 sub-categories) this benchmark minimizes the likelihood of overfitting. Task accuracy is calculated based on whether the chosen response receives a higher reward than the rejected response. We directly evaluate their open dataset reward-bench \footnote{\url{https://huggingface.co/datasets/allenai/reward-bench}} 



\section{Hyper parameter tuning} \label{sec:hypertuning}

We performed hyperparameter tuning for the \textbf{GazeConcat} method and the baseline, and we replicated them in the \textbf{GazeAdd} method, as testing all combinations is computationally very expensive. For each dataset, 15\% is reserved for validation to perform hyperparameter tuning. The best-performing checkpoints are selected based on the lowest validation loss and are subsequently used for performance evaluation in the test split. We trained for two epochs, as described in \cite{wang_helpsteer2_2024} and in line with trends observed in \cite{touvron_llama_2023} where they found that training longer can lead to over-fitting. We perform a grid search for the optimal batch size, testing \{8, 16, 32\} values. AdamW optimizer \citep{loshchilov_decoupled_2019} is used and the learning rate is tuned within the range of \{1, 5, 10, 50\}e-6, inspired by the values reported in \cite{touvron_llama_2023, cui_ultrafeedback_2024, wang_helpsteer2_2024}. Additionally, we explored different scheduler configuration, comparing constant, linear, and cosine with a minimum learning rate.

\begin{figure}[tb!]
    \begin{center}
    \includegraphics[width=0.5\linewidth]{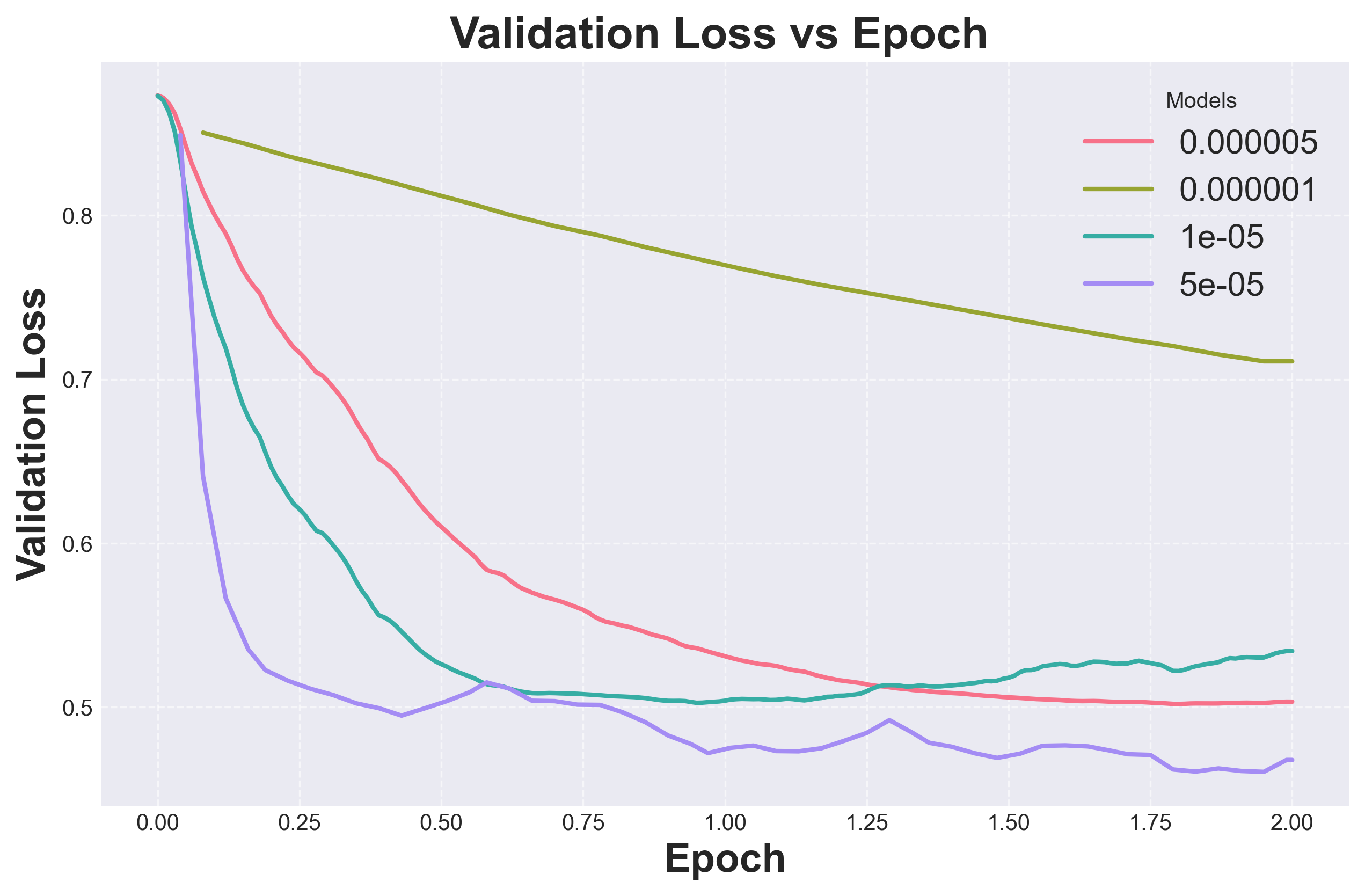}
    \end{center}
    \caption{Validation loss with different LR on ConcatReward, batch size 8, features: $f1$ and Meta-Llama-3-8B-Instruct base model}
    \label{fig:hyper_lr}
\end{figure}

\begin{figure}[tb!]
    \begin{center}
    \includegraphics[width=0.5\linewidth]{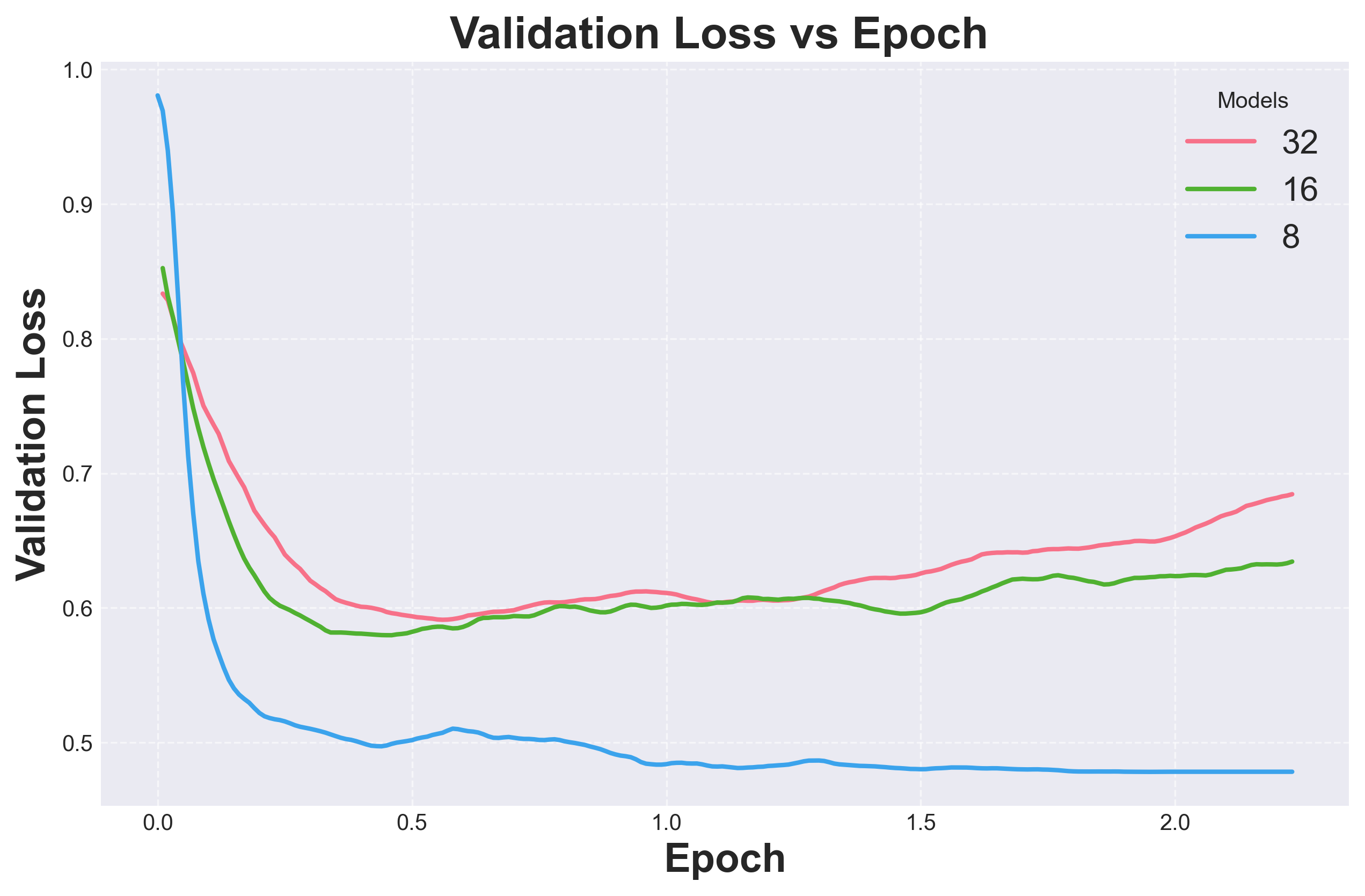}
    \end{center}
    \caption{Validation loss with different batch size, learning rate: 5e-5, features: $f1$ and Meta-Llama-3-8B-Instruct base model}
    \label{fig:hyper_batch}
\end{figure}

\begin{figure}[tb!]
    \centering
    \begin{subfigure}{0.45\linewidth}
        \centering
        \includegraphics[width=\linewidth]{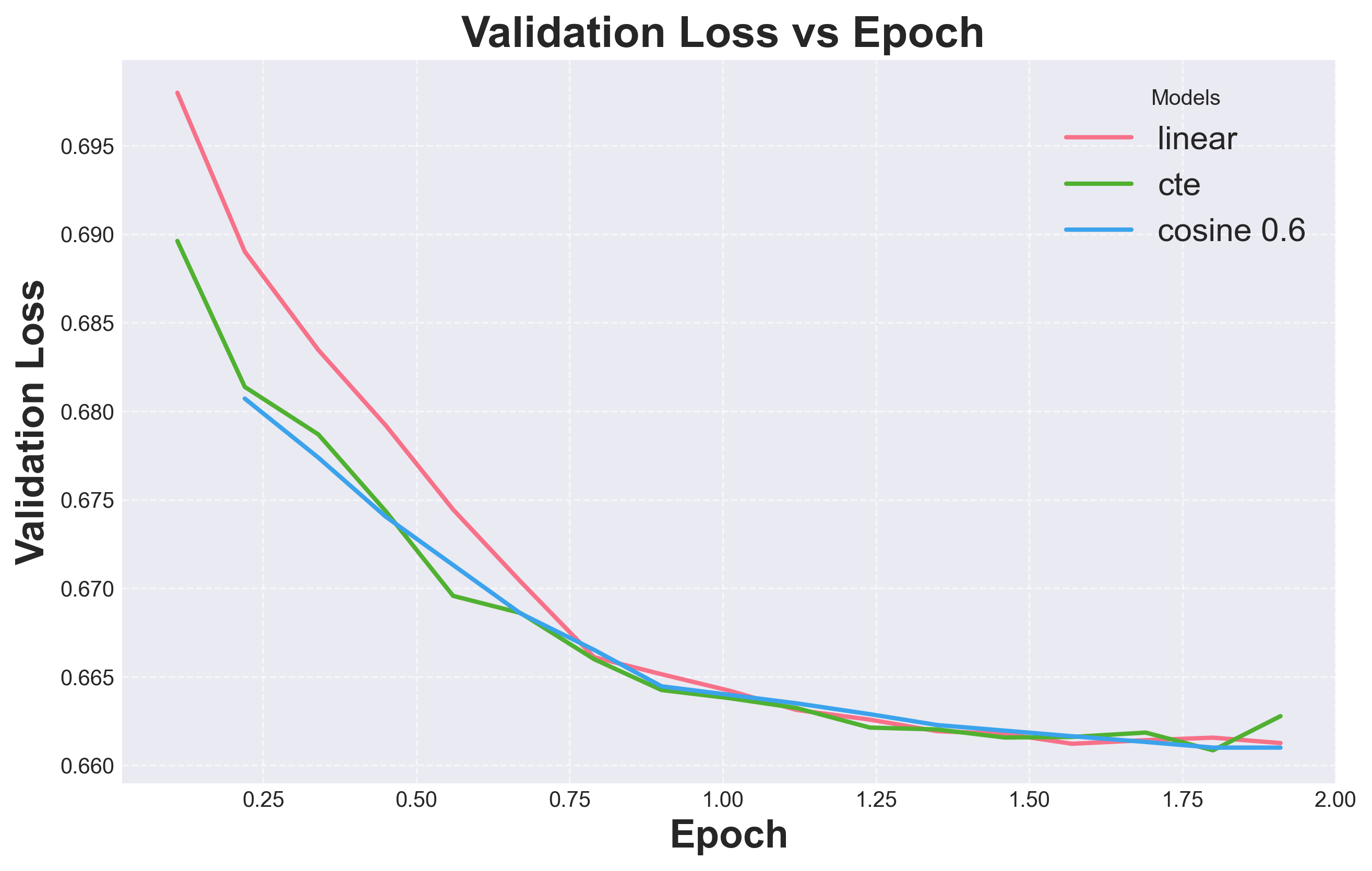}
        \caption{Validation Loss Comparison for different LR Schedulers }
        \label{scheduler_a1_plot}
    \end{subfigure}
    \hfill 
    \begin{subfigure}{0.45\linewidth}
        \centering
        \includegraphics[width=\linewidth]{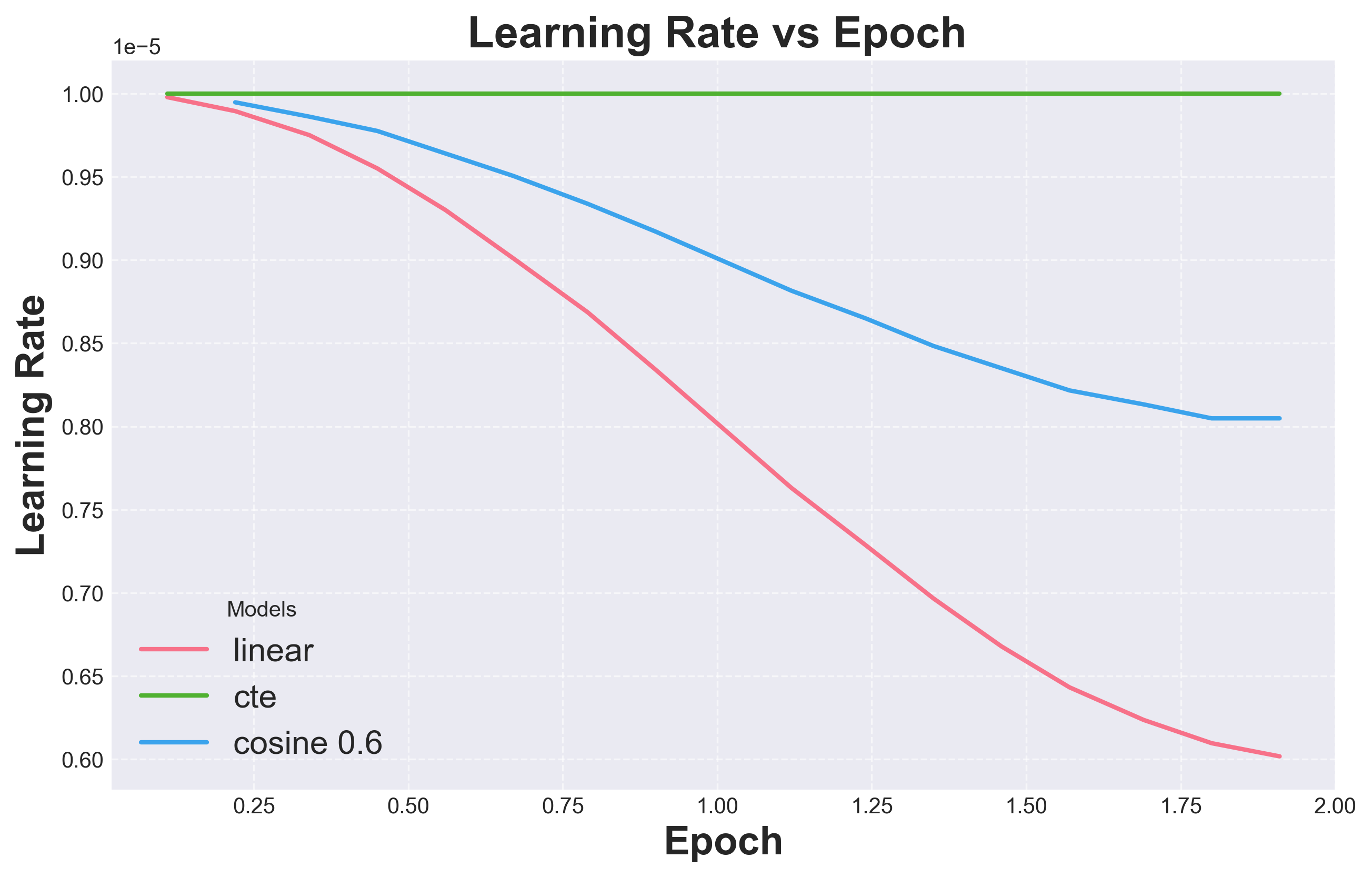} 
        \caption{Learning rate comparison for different LR Schedulers}
        \label{scheduler_a2_plot}
    \end{subfigure}
    \caption{ConcatReward, LR: 0.00001, features: $f1$ and Meta-Llama-3-8B-Instruct base model}
    \label{fig:hyper_lr2}
\end{figure}

\begin{figure}[tb!]
    \centering
    \begin{subfigure}{0.45\linewidth}
        \centering
        \includegraphics[width=\linewidth]{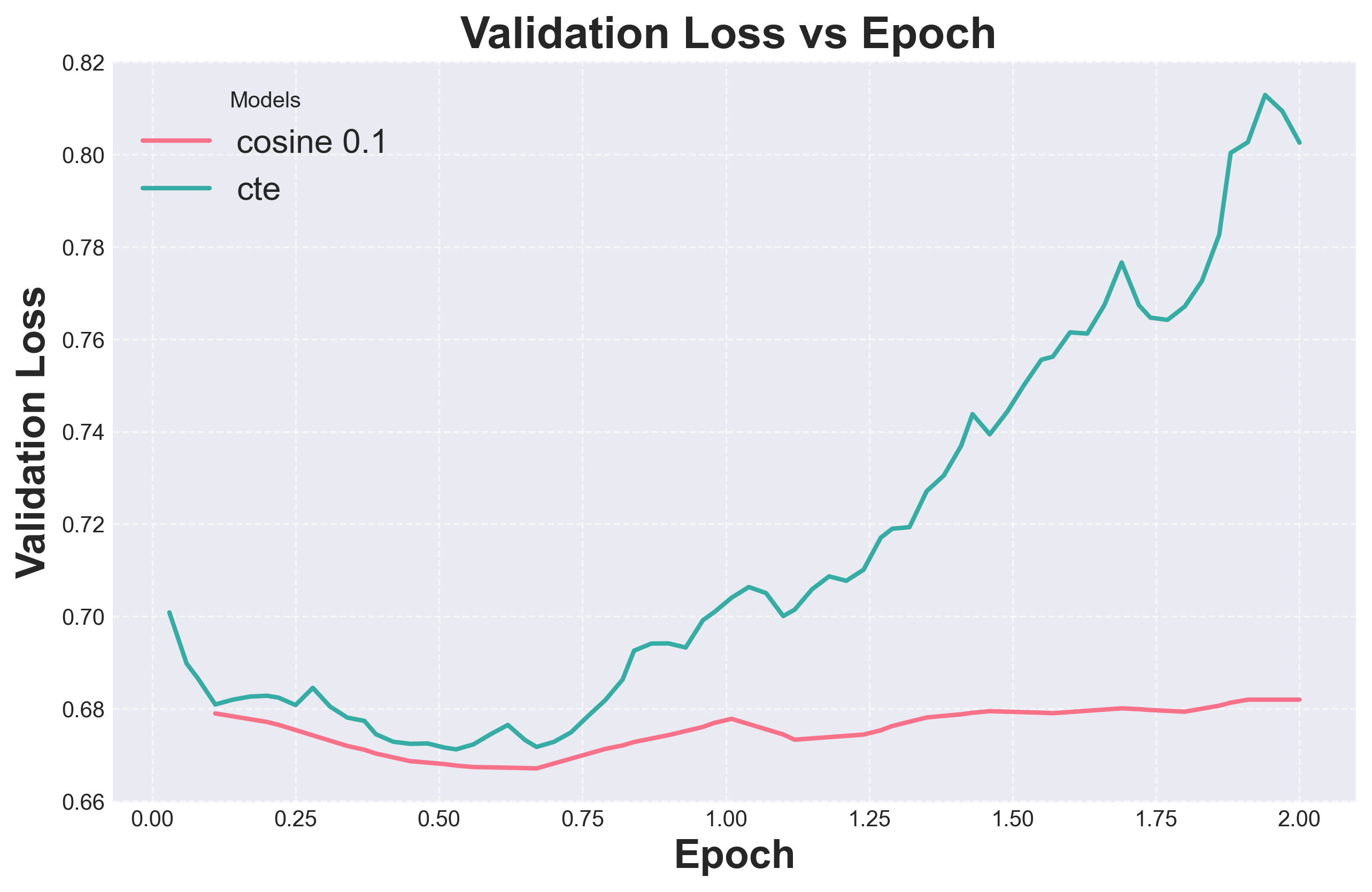}
        \caption{Validation Loss Comparison for different LR Schedulers}
        \label{scheduler_b1_plot}
    \end{subfigure}
    \hfill 
    \begin{subfigure}{0.45\linewidth}
        \centering
        \includegraphics[width=\linewidth]{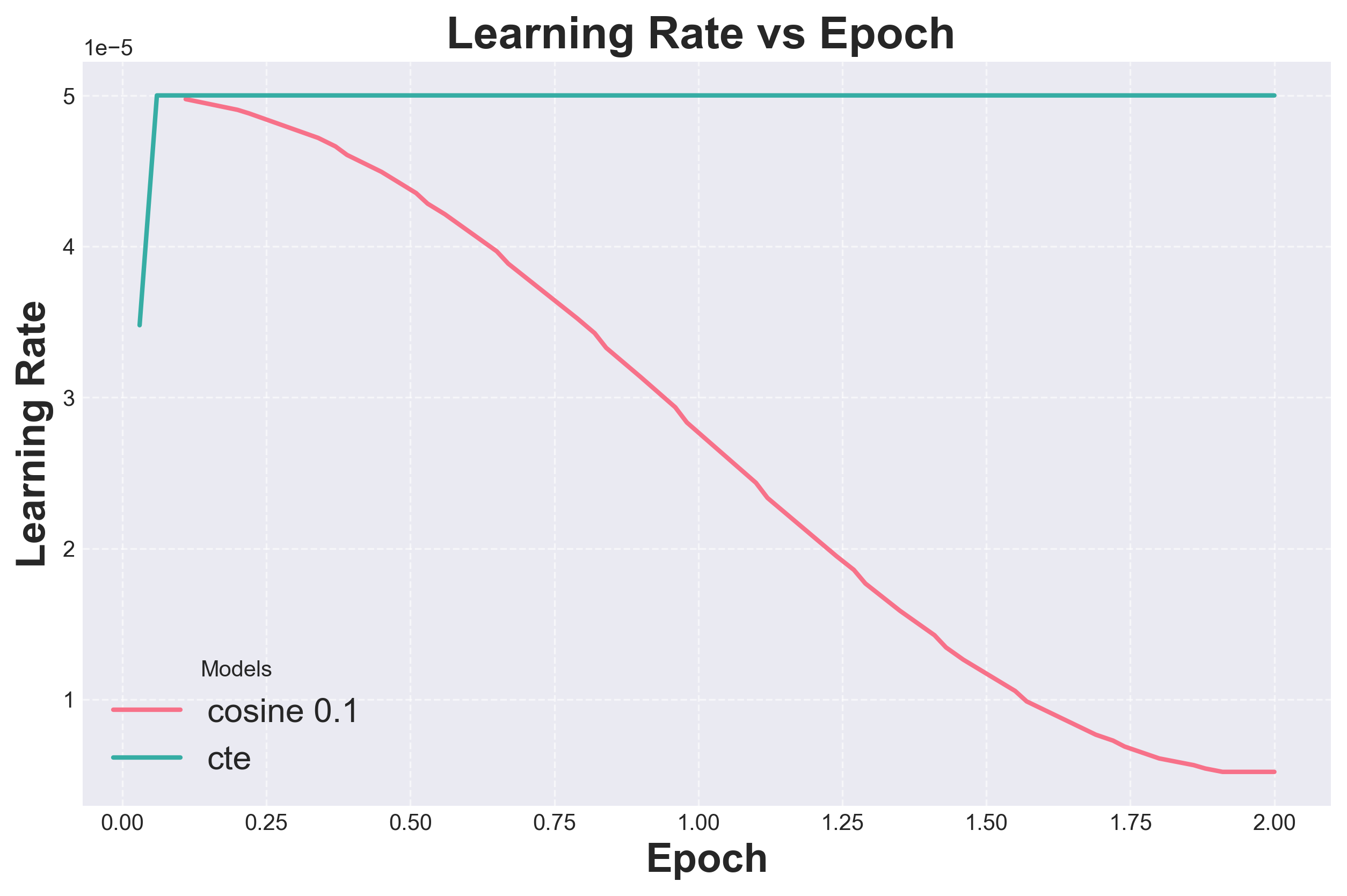} 
        \caption{Learning rate comparison for different LR Schedulers}
         \label{scheduler_b2_plot}
    \end{subfigure}
    \caption{ConcatReward, LR: 0.00005, features: $f1$ and Meta-Llama-3-8B-Instruct base model}
    \label{fig:hyper_lr3}
\end{figure}

In general, the parameter that most affected the validation results was the learning rate. For the others, we ended up choosing values that worked well across all combinations. We achieved better results in both the baseline and the models concatenating the \acrshort{et} features with a learning rate of 0.0005 ( \autoref{fig:hyper_lr}). In \autoref{scheduler_a1_plot} and \autoref{scheduler_a2_plot}, the validation loss and learning rate with different schedulers are shown. For lower learning rates, such as 0.00001, the scheduler had little effect. However, with higher learning rates, using a scheduler helped to mitigate overfiting. In \autoref{scheduler_b1_plot} and \autoref{scheduler_b2_plot}, it represents validation loss and learning rates with a higher learning rate of 0.0005. We opted to use this 0.00005 learning rate for all experiments, employing a cosine learning rate scheduler with a minimum learning rate of 0.7, in line with other studies such as \cite{wang_helpsteer2_2024, touvron_llama_2023}.  We did not find a significant effect of training batch size on validation accuracy, but we opted for a value of 8, which often (especially with high learning rates and without a scheduler) was the one that tended to overfit the least ( \autoref{fig:hyper_batch}).

\textbf{\acrshort{et} features projector}
The PyTorch architecture of our \acrshort{et} features projector is shown below. $num\_features$ varies between 1, 5, and 2 (depending on the configuration used $fcomb_{1}$, $fcomb_{2.5}$, $fcomb_{2.2}$  \autoref{sec:results}). $p_{1}, p_{2}$ are dropout values. Finally, after testing different combinations, we used $0.1$ and $0.3$.  This model has 0.53M parameters.
\\
\\
\\
\\
\\
\\
\\
\\
\\
\\
\\
\\
\\
\begin{lstlisting}[style=pytorch, caption={PyTorch architecture of our gaze features projector}]
self.fixations_embedding_projector = nn.Sequential(
    nn.Linear(num_features, 128),
    nn.LayerNorm(128),
    nn.ReLU(),
    nn.Dropout(p=p_1),
    nn.Linear(128, hidden_size),
    nn.Dropout(p=p_2),
)
\end{lstlisting}

\end{document}